\documentclass{article}
\usepackage{iclr2027_conference,times}
\usepackage{amsmath,amsfonts,bm}

\def\eqref#1{equation~\ref{#1}}

\def\1{\bm{1}}

\DeclareMathAlphabet{\mathsfit}{\encodingdefault}{\sfdefault}{m}{sl}
\SetMathAlphabet{\mathsfit}{bold}{\encodingdefault}{\sfdefault}{bx}{n}

\usepackage{hyperref}
\usepackage{url}

\usepackage[utf8]{inputenc}
\usepackage[T1]{fontenc}
\usepackage{microtype}
\usepackage{amsmath,amssymb,amsthm}
\usepackage{bbm}
\usepackage{nicefrac}
\usepackage{graphicx}
\usepackage{booktabs,array}
\usepackage{xcolor}
\usepackage{capt-of}
\usepackage{enumitem}
\usepackage{wrapfig}
\usepackage{float}
\usepackage{pifont}
\usepackage{forest}
\usepackage{longtable}
\usepackage{tabularx}
\usepackage{placeins,needspace}
\usepackage[skins]{tcolorbox}
\usepackage[scaled=.95,varl,narrow]{inconsolata}
\usepackage{caption}
\usepackage{colortbl}
\newcommand{\stage}[2]{\textcolor{#1}{\textbf{#2}}}

\definecolor{dred}{RGB}{153,80,43}
\definecolor{dblue}{RGB}{0,114,178}
\definecolor{frankcrimson}{HTML}{B23C67}
\definecolor{winpurple}{HTML}{5B348E}
\definecolor{winorange}{HTML}{BA5F08}
\definecolor{birdgenerate}{HTML}{0072B2}
\definecolor{birdverify}{HTML}{B07800}
\definecolor{birdtrain}{HTML}{23836E}
\definecolor{birdevaluate}{HTML}{9B5D98}
\definecolor{birdselect}{HTML}{CE6237}
\definecolor{birdupdate}{HTML}{B23C67}
\definecolor{birdframe}{HTML}{C8D2DB}
\definecolor{birdback}{HTML}{FBFCFD}
\definecolor{birdhead}{HTML}{EAF0F5}
\definecolor{antclay}{HTML}{D97757}
\definecolor{antslate}{HTML}{3D3D3A}
\definecolor{antcloud}{HTML}{BFBFBA}

\newcommand{\yes}{\textcolor{antslate}{\ding{51}}}
\newcommand{\add}{\textcolor{antclay}{\ding{52}}}
\newcommand{\no}{\textcolor{antcloud}{--}}
\hypersetup{
    colorlinks=true,
    breaklinks=true,
    urlcolor=dblue,
    linkcolor=dred,
    citecolor=dblue
}

\usepackage[capitalize,noabbrev,nameinlink]{cleveref}
\creflabelformat{equation}{#2#1#3}
\crefname{appendix}{Appendix}{Appendices}
\Crefname{appendix}{Appendix}{Appendices}

\title{A Bird's-Eye View of Iterative Reward Design}

\author{%
  Logan~M.~Bhamidipaty$^{1,2}$\thanks{Equal contribution. Correspondence to \texttt{l.m.bhamidipaty@sms.ed.ac.uk}.} \quad
  Lauren~Robson$^{1}$\footnotemark[1] \\
  \textbf{Linda~Petrini$^{3}$ \quad Shengrui~Lyu$^{3}$ \quad Kamal~Ndousse$^{3}$} \\[3pt]
  $^{1}$Anthropic Fellows Program \quad $^{2}$University of Edinburgh \quad $^{3}$Anthropic
}

\iclrfinalcopy
\begin{document}

\maketitle
\lhead{Preprint}

\begin{abstract}
Designing effective reward functions in RL typically requires substantial expertise and trial and error. Recent work automates this process with LLM-based systems that generate and iteratively improve reward code using policy feedback. However, these methods are often hard to compare because they differ in implementation details, feedback assumptions, and evaluation environments. To address this, we introduce a \textbf{B}enchmark for \textbf{I}terative \textbf{R}eward \textbf{D}esign (\textbf{BIRD}) that expresses existing methods in a unified configuration and evaluation space.\footnote{\url{https://github.com/safety-research/bird}} This lets us compare algorithms directly, ablate individual design choices, and prototype new components under matched feedback conditions and policy-training budgets. Across MuJoCo, Meta-World, Assistax, and HumanoidBench, we identify a small set of simple design choices that consistently improve performance. Combining these choices yields significantly better performance than the evaluated methods from prior work. Our results highlight the strength of simple baselines and motivate further study of when additional algorithmic complexity improves iterative reward design. 

\end{abstract}

\section{Introduction}

A central difficulty in reinforcement learning (RL) is designing reward functions that support efficient learning while remaining aligned with designer intent. At one extreme, a reward may be easy to optimize but poorly aligned: shaping terms and proxy rewards can improve sample efficiency but may bias the objective \citep{ng1999policy, pan2022effects} or induce reward hacking \citep{amodei2016concrete, skalse2022defining}. At the other extreme, a reward may be well aligned but difficult to optimize: binary completion flags in robotics, for example, can be too sparse to support efficient policy learning without heuristics \citep{andrychowicz2018hindsight}. Navigating this tradeoff often requires trial and error, which is unsurprisingly both laborious and error-prone even for expert designers \citep{booth2023perils}.

\begin{figure}[t]
  \centering
  \includegraphics[width=\linewidth]{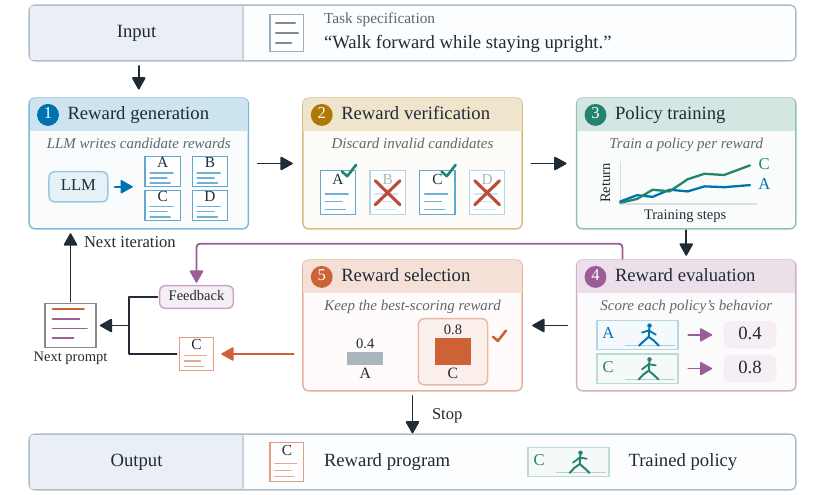}
  \caption{Most iterative reward design (IRD) methods follow this basic loop. BIRD uses this shared structure to represent methods in a common format, making them easier to compare.}
  \label{fig:bird-loop}
\end{figure}

This process, known as \emph{iterative reward design} (IRD), follows a basic loop: propose a reward, train a policy, evaluate its behavior, revise the reward, and repeat as needed (\cref{fig:bird-loop}).\footnote{Sometimes humorously referred to as ``graduate student descent.''} To reduce the burden on human designers, methods like Eureka \citep{ma2024eureka} and Text2Reward \citep{xie2024text2reward} have proposed using LLMs to automate reward generation and revision. These rewards typically far surpass human-written ones and have thus inspired a large body of work on LLM-assisted IRD \citep{hazra2025revolve, gao2025rfagent, ma2025gran, lee2026rda, jaswal2026limen}.

The growing literature offers a broad menu for generating, evaluating, and updating reward functions, but provides limited guidance about which design choices work reliably. Within papers, benchmarks and ablations typically test design choices within particular algorithms and environments, leaving unclear whether those choices also improve other algorithms, remain effective in other environments, or retain their benefits when combined. Across papers, differences in feedback assumptions, harness implementations, model backbones, and policy-training procedures make it difficult to compare methods fairly or isolate the effects of individual choices. Moreover, higher benchmark performance does not by itself establish better alignment with the intended objectives. These limitations leave four central questions unresolved: (1) \textit{Which IRD methods perform best across tasks?}; (2) \textit{Which design choices drive performance?}; (3) \textit{Can useful design choices be combined into better methods?}; and (4) \textit{Does better performance mean better alignment?}

To address these questions, we develop a \textbf{B}enchmark for \textbf{I}terative \textbf{R}eward \textbf{D}esign (\textbf{BIRD}), built around a shared, configurable implementation of the IRD loop. BIRD represents each method as a configuration specifying how to generate candidate rewards, train and evaluate policies, and select and revise rewards. When methods share a design choice, they reuse its implementation. Researchers can therefore reproduce existing methods, modify individual choices, or combine choices from different methods within the same framework. BIRD pairs this implementation with a common evaluation protocol, enabling comparisons under matched feedback conditions and policy-training budgets.

Our contributions are fourfold.
\begin{enumerate}[leftmargin=*, topsep=2pt, itemsep=2pt, parsep=0pt, partopsep=0pt]
    \item We introduce BIRD (\cref{sec:bird}), a unified framework for implementing, comparing, ablating, and developing IRD methods, with implementations for 14 published methods (\cref{app:methods}) and evaluation procedures for MuJoCo \citep{todorov2012mujoco}, Meta-World \citep{yu2019metaworld}, Assistax \citep{hinckeldey2025assistax}, and HumanoidBench \citep{sferrazza2024humanoidbench}.

    \item We compare 6 existing methods (\cref{sec:results-methods}) and ablate 30 design choices (\cref{sec:results-ablation}) under matched experimental conditions. We find that many design choices, both from the literature and our own proposals, produce small or inconclusive effects when deployed across new tasks and algorithmic stacks, and that only a small set of simple design choices clearly helps. This suggests that intuition-driven design choices often fail to transfer across tasks and methods, and that systematic empirical benchmarking is a necessary complement.

    \item Using BIRD, we develop two novel IRD algorithms ERA-U and ERA-S (unsupervised and supervised variants) that achieve significantly better performance over established baselines in their respective categories on the environment suites (\cref{sec:results-recipe}).

    \item To preliminarily assess IRD alignment, we conduct a small human study $(n = 2)$. Consistent with prior work on specification gaming \citep{krakovna2020specification}, we find that rewards designed by IRD methods can induce policies that score highly on benchmarks through unnatural or even harmful behavior (\cref{sec:results-human-eval}).
\end{enumerate}

\section{Preliminaries}
\label{sec:prelims}

We assume most readers have a passing familiarity with RL and so omit the usual MDP and policy definition regurgitation; interested readers and those new to the field can consult \citet{sutton1998reinforcement}. Instead, we focus on \textit{what} iterative reward design is and \textit{how} the process works.\footnote{We have written a ``truly marvelous [prior work], which this margin is too narrow to contain'' (\Cref{app:prior-work}).}

\begin{quote}
\textit{Iterative reward design (IRD) is the process of designing rewards by trial and error based on policy feedback, often using LLMs or VLMs.}
\end{quote}

\noindent Before walking through the stages, we strongly encourage readers to skim the terminology table in \Cref{app:terminology}, which clarifies frequently reused terms such as ``design choice'' and ``reference reward.'' As illustrated in \Cref{fig:bird-loop}, IRD algorithms can typically be decomposed into the following stages.

\begin{figure}[t]
  \centering
  \includegraphics[width=\linewidth]{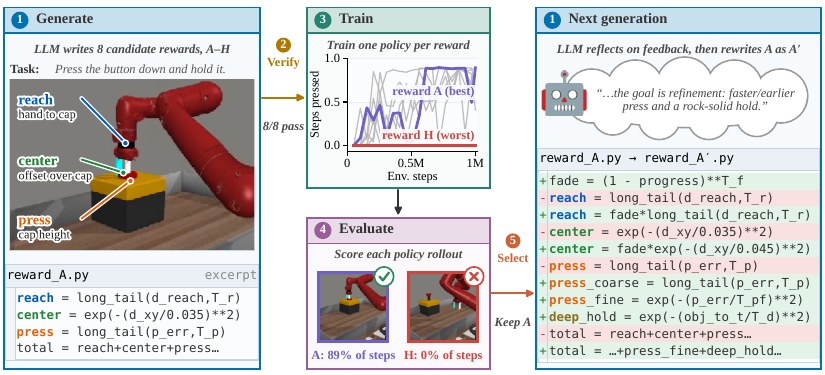}
    \caption{\textbf{One iteration of Eureka on Meta-World's button-press task.}
    The LLM generates eight reward functions, all of which pass verification.
    Each reward is used to train a PPO policy for 1M environment steps.
    Reward $A$ produces the best policy (89\% task success), while $H$ produces
    the worst (0\%). Eureka greedily retains $A$ and provides the LLM with
feedback on how its reward components (e.g., \texttt{reach},
\texttt{press}, and \texttt{center}) evolve during training.
The LLM uses this feedback to revise $A$ into $A'$.
All results are from an actual run.}
  \label{fig:eureka-worked-example}
\end{figure}

\textbf{1.\ Reward generation.} The process starts with an LLM or VLM proposing candidate rewards from a task description and information about the environment, often its source code. Any reward shipped with the environment (its \emph{reference reward}) is withheld, so the model designs rewards from scratch. Rewards are usually represented as Python functions (see \Cref{fig:eureka-worked-example}). On the first iteration, the model works only from what it is given about the task; on later iterations, it can also draw on feedback from earlier attempts.

\textbf{2.\ Reward verification.} Training a policy under every generated reward can get expensive, so some methods first screen candidates to decide which are worth training on. Checking for coding errors is particularly useful when using weaker models. Other checks try to estimate reward quality before training a policy. CARD, for example, retains a buffer of trajectories from earlier iterations and validates candidate rewards by checking whether they rank successful trajectories above failed ones \citep{sun2025card}.

\textbf{3.\ Policy training.} The surviving candidates then move on to policy training, which is essentially vanilla RL. One or more policies are trained under each reward function. The design choices at this stage are fairly standard: which RL algorithm to use (e.g., PPO or SAC), how many environment steps to train for, and whether to train from scratch or warm-start from an earlier policy.

\textbf{4.\ Reward evaluation.} In the next stage, methods determine how well each reward worked. Here methods split into two kinds. \textit{Supervised} IRD methods use the benchmark score, which grades trained policies against the task (\Cref{app:terminology}); Eureka, for example, combines it with statistics about reward components over the course of training \citep{ma2024eureka}. \textit{Unsupervised} methods never access the benchmark score: RDA, for example, replaces it with a VLM that rates task completion from image frames sampled uniformly across a policy rollout \citep{lee2026rda}.

\textbf{5.\ Reward selection.} After curating information about each candidate, methods then select a winner or a set of elite candidates. A common design choice is to greedily keep the highest-scoring candidate, as in Eureka and RDA \citep{ma2024eureka,lee2026rda}. Other methods use populations or search trees to keep more candidates in play \citep{hazra2025revolve,gao2025rfagent}. The comparison itself can also be based on preferences: REvolve uses human preferences \citep{hazra2025revolve}, while Gran Turismo\footnote{\citet{ma2025gran} do not name their method, so we refer to it by its environment: Gran Turismo.} uses VLM preferences and then collects human feedback on the selected policy \citep{ma2025gran}. These choices determine which rewards become the starting point for the next round of generation.

\section{BIRD}
\label{sec:bird}

As we explain in the introduction, IRD methods are hard to compare because they typically have different implementations, backbones, and assumptions. The key insight of BIRD is that these difficulties can be largely overcome by leveraging the shared structure of the IRD loop (\cref{sec:prelims}).

\begin{figure}[t]
  \centering
  \begin{minipage}[c]{0.52\textwidth}
    \centering
    \includegraphics[width=\linewidth]{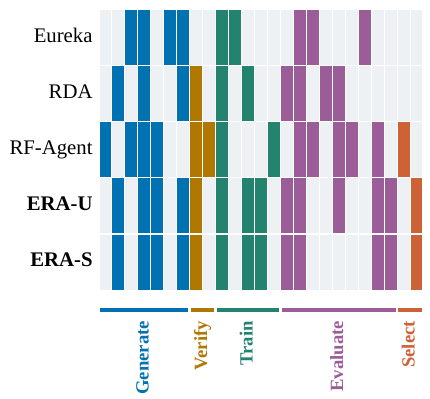}
  \end{minipage}\hfill
  \begin{minipage}[c]{0.45\textwidth}
    \begin{tcolorbox}[enhanced, width=\linewidth, arc=3pt, boxrule=0.5pt,
      colframe=birdframe, colback=birdback, colbacktitle=birdhead, coltitle=black,
      fonttitle=\normalsize\fontfamily{zi4}\selectfont, title=eureka.yaml,
      left=5pt, right=5pt, top=4pt, bottom=4pt, toptitle=3pt, bottomtitle=3pt]
    \normalsize\fontfamily{zi4}\selectfont\frenchspacing\raggedright\setlength{\parindent}{0pt}\setlength{\parskip}{0pt}
    \mbox{\textcolor{birdgenerate}{generate:}}\par
    \mbox{\phantom{xx}n\_candidates: 8}\par
    \mbox{\phantom{xx}history\_mode: last\_iteration}\par
    \mbox{\textcolor{birdverify}{verify:}}\par
    \mbox{\phantom{xx}enabled: false}\par
    \mbox{\textcolor{birdtrain}{train:}}\par
    \mbox{\phantom{xx}init: from\_scratch}\par
    \mbox{\textcolor{birdevaluate}{evaluate:}}\par
    \mbox{\phantom{xx}fitness:}\par
    \mbox{\phantom{xxxx}source: native}\par
    \mbox{\textcolor{birdselect}{select:}}\par
    \mbox{\phantom{xx}rule: argmax\_fitness}\par
    \end{tcolorbox}
  \end{minipage}

\caption{\textbf{BIRD represents IRD methods as configurations.} \textit{Left:} Each row is a method and each column is a design choice key, grouped by the stages from \Cref{sec:prelims}. Gray cells represent design choices set by the BIRD default (e.g., \texttt{init=from\_scratch}); colored cells are design choices set by the algorithm (e.g., Eureka's \texttt{n\_candidates=8}). As we explain in \Cref{sec:results-recipe}, ERA-U and ERA-S use the RDA configuration as a backbone. \textit{Right:} Each configuration is a short YAML file. An excerpt from Eureka's BIRD configuration.}  \label{fig:bird-config}
\end{figure}

Following Unifloral \citep{jackson2025unifloral}, we represent IRD methods as ``configurations.'' Each configuration specifies the choices made at each IRD stage: which feedback to collect, how to select candidates, and so on (\Cref{fig:bird-config}). When methods share a choice, they reuse its implementation. For example, Eureka and RDA use different feedback sources but make the same selection choice. Their BIRD configurations therefore specify different scoring signals but use the same ranking code. This helps prevent implementation differences (e.g., differences in tie-breaking logic) from becoming confounders.

BIRD currently represents 14 IRD methods (\cref{app:methods}). To  the best of our knowledge, our implementations are correct. We verify our implementations against the original papers and reference source code; however, not all papers open-source their code, and some model backbones have been deprecated, making complete fidelity difficult, if not impossible.

\section{Experimental Results}
\label{sec:results}

We use BIRD to investigate four key questions.

\begin{enumerate}[leftmargin=*, topsep=2pt, itemsep=2pt, parsep=0pt, partopsep=0pt]
    \item \textbf{Which methods work best?}
    We compare methods from prior work (\cref{app:methods}) across MuJoCo \citep{todorov2012mujoco}, Meta-World MT10 \citep{yu2019metaworld}, Assistax \citep{hinckeldey2025assistax}, and HumanoidBench \citep{sferrazza2024humanoidbench} under matched policy-training budgets.

    \item \textbf{Why do those methods work?} We isolate individual design choices to distinguish broadly useful
    components from improvements that depend on a particular task or
    configuration.

    \item \textbf{Can we combine useful choices?}
    We evaluate whether a small set of simple design choices provides an
    effective recipe for reward design.

    \item \textbf{What does doing well mean?}
    We examine the alignment between benchmark scores and human
    judgments of whether the task is accomplished and how it is performed.
\end{enumerate}

\subsection{Comparing IRD Methods}
\label{sec:results-methods}

\begin{figure}[t]
  \centering
  \includegraphics[width=\linewidth]{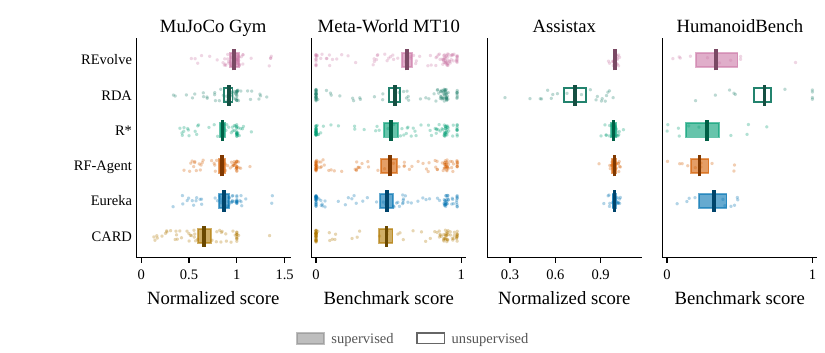}
  \caption{\textbf{Comparing IRD methods across four environment suites.} Each seed represents one run on an environment. Ticks show mean across tasks and seeds. Bars in this plot and all others are 95\% BCa bootstrap confidence intervals. Experimental details are in \cref{app:experimental-details}.}
\label{fig:final-performance}
\end{figure}

We first use BIRD to compare 6 published IRD methods. We find that REvolve performs the strongest across all suites, only falling behind RDA in HumanoidBench; no other method consistently outperforms any other with non-overlapping CIs, suggesting these methods are surprisingly similar performance-wise. We provide per-task performance and compute efficiency plots in \Cref{app:performance,app:compute}.

As we elaborate in \Cref{sec:results-human-eval}, we find RDA---the only unsupervised method---performs more inconsistently on Assistax and HumanoidBench. We hypothesize that this reflects the difficulty of assessing task completion from visual feedback alone. For example, determining whether a button has been pressed is relatively straightforward, whereas determining whether a robot has successfully bathed a bedridden patient may require information that is difficult to infer purely visually.

We also observe diminishing returns from additional iterations. This is likely because most Python reward functions are represented as weighted sums and subsequent iterations mainly tune coefficients rather than introduce large structural changes (see excerpt in \cref{fig:eureka-worked-example}). For that reason, some methods like RF-Agent add countermeasures against this kind of incremental iteration \citep{gao2025rfagent}. For further evidence, see the EPIC reward distance \citep{gleave2021quantifying} analysis in \Cref{app:epic}.

\subsection{Ablating IRD Design Choices}
\label{sec:results-ablation}

\begin{figure}[t]
  \centering
  \includegraphics[width=\linewidth]{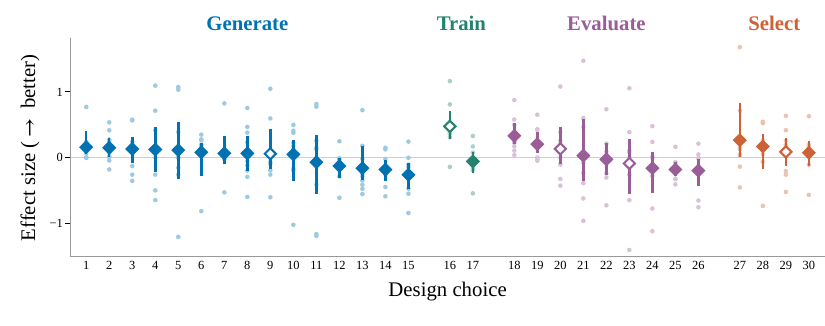}
  \caption{\textbf{Most design choices show no clear benefit.}
Effects of enabling 30 design choices across applicable baselines
(Eureka, RDA, CARD) and nine tasks. Hollow diamonds mark choices introduced in BIRD. Only four choices clearly help: (16) peak own-reward checkpoint, (18)
summarized feedback, (19) task-success fitness, and (1) four-turn
history.
See \cref{app:design-choices} for all choices and experimental details.}
  \label{fig:ablation}
\end{figure}

The natural next direction after comparing IRD methods is investigating which design choices drive performance. To investigate, we ablate 30 choices drawn from existing algorithms and our own proposals. We select these manually, as exhaustively evaluating BIRD's more than 300 configurable choices is computationally infeasible.

We find that most choices do not consistently improve performance over the baseline, including many of our own proposals (\cref{fig:ablation}). Several of the strongest choices are simple, such as greedily retaining checkpoints and summarizing feedback. Together, these results suggest that additional algorithmic complexity does not reliably improve IRD.

The variability across tasks may partly explain why promising choices do not produce consistent gains. More broadly, the difficulty of manual reward design \citep{booth2023perils} gives reason to question whether human intuitions about reward revision reliably translate into effective automated methods.

\subsection{A Simple Recipe for IRD}
\label{sec:results-recipe}

\begin{figure}[t]
  \centering
  \begin{minipage}[t]{0.46\linewidth}
    \vspace{0pt}\centering
    \includegraphics[width=\linewidth]{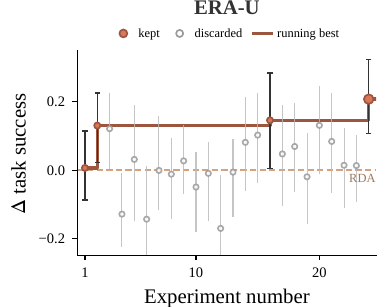}
    \captionof{figure}{\textbf{Hill-climbing unsupervised IRD.} We tried 24 configurations on
    four unsaturated MT10 tasks and stopped at $+0.21$ benchmark score over RDA without trying to
    climb further. Bars: 95\% BCa intervals over 3-7 seeds per experiment. Full details in \cref{app:exp-hillclimb}.}
    \label{fig:hill-climb-unsupervised}
  \end{minipage}\hfill
  \begin{minipage}[t]{0.51\linewidth}
    \vspace{0pt}\centering
    \setlength{\abovecaptionskip}{0pt}
    \captionof{table}{\textbf{The ERA recipes} (U: ERA-U, S: ERA-S). \add{} added to RDA;
    \yes{} inherited. ERA-S and ERA-U differ in exactly one key, the source of the signal used to rank candidates (shaded).}
    \label{tab:era-recipes}
    \vspace{11pt}
    \footnotesize
    \setlength{\tabcolsep}{2.2pt}
    \renewcommand{\arraystretch}{1.05}
    \begin{tabular}{l l c cc}
      \toprule
      & & & \multicolumn{2}{c}{\textbf{Ours}} \\
      \cmidrule(l){4-5}
      & Design choice & RDA & \textcolor{antclay}{\textbf{U}} & \textcolor{antclay}{\textbf{S}} \\
      \midrule
      \stage{birdgenerate}{Gen.}  & Task decomposition           & \yes & \yes & \yes \\
                                  & Design thought (RF-Agent)    & \no  & \add & \add \\
      \addlinespace[2pt]
      \stage{birdtrain}{Train}    & Warm start                   & \yes & \yes & \yes \\
                                  & Peak own-reward ckpt.        & \no  & \add & \add \\
      \addlinespace[2pt]
      \stage{birdevaluate}{Eval.} & Per-subtask feedback         & \yes & \yes & \yes \\
                                  & Continuous score, $\times2$  & \no  & \add & \add \\
                                  & States shown to judge        & \no  & \add & \add \\
      \rowcolor{antclay!12}       & Selection: VLM completion rating           & \yes & \yes & \no  \\
      \rowcolor{antclay!12}       & Selection: reference reward           & \no  & \no  & \add \\
      \addlinespace[2pt]
      \stage{birdselect}{Sel.}    & Greedy selection             & \yes & \yes & \yes \\
                                  & Keep incumbent unless beaten & \no  & \add & \add \\
      \bottomrule
    \end{tabular}
  \end{minipage}
\end{figure}

The ablations in \cref{sec:results-ablation} suggest that the choices that help reliably tend to be simple. We therefore asked whether stacking a few simple choices on an existing method is enough to outperform the methods we compared. Starting from the RDA configuration, we hill-climbed on four unsaturated Meta-World MT10 tasks (pick-place, push, peg-insert-side, and drawer-open), changing one design choice at a time. To keep the search cheap, most configurations ran at a reduced budget of 24--30 policy trainings per search. We kept the search deliberately small and stopped after 24 configurations, when the last one, re-run at the full budget of 40 trainings, led RDA by more than 0.2 in benchmark score (\cref{fig:hill-climb-unsupervised}). The result, ERA-U, adds five choices to RDA (\cref{tab:era-recipes}), including the peak own-reward checkpoint that helped most clearly in \cref{fig:ablation}.

\begin{figure}[t]
  \centering
  \includegraphics[width=\linewidth]{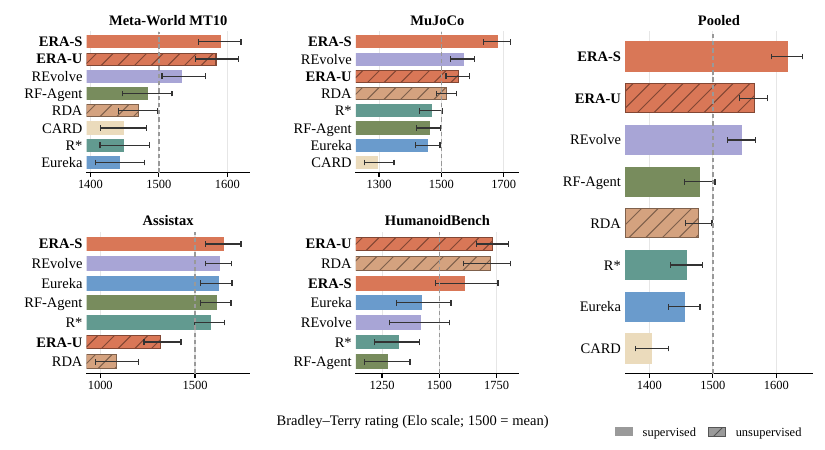}
  \caption{\textbf{ERA-S and ERA-U rank first and second overall.} Bradley--Terry ratings on the Elo scale (mean $1500$, dashed line) for each benchmark and pooled over all 21 tasks. Within each task, every seed of one method is compared with every seed of another; the higher benchmark score wins and ties count as half a win. Intervals resample seeds with the task set held fixed. Hatched bars mark unsupervised methods. CARD is run only on MuJoCo and MT10.}
  \label{fig:bt-rankings}
\end{figure}

To obtain a supervised recipe, we did no further search. ERA-S is ERA-U with a single key flipped: the one that sets the ranking signal, from the VLM completion rating to the reference reward (shaded rows of \cref{tab:era-recipes}).

Although ERA-U was hill-climbed on only four MT10 tasks and ERA-S was not tuned at all, both recipes transfer well. Under our matched protocol, they take first and second place in the pooled Bradley--Terry ranking (\cref{fig:bt-rankings}): ERA-S outperforms REvolve, the strongest supervised baseline, and ERA-U outperforms RDA, its unsupervised parent, both with non-overlapping CIs. The ranking holds when the four development tasks are excluded, and when all seven tasks used at any point during development are excluded (\cref{app:exp-heldout}). ERA-U, which never sees the benchmark signal, also ranks above every supervised method except REvolve. As we discuss in \cref{sec:discussion}, we present ERA-U and ERA-S not as methods to adopt in their own right (we expect further search to surpass them substantially), but as evidence that simple choices and a small hill-climb go a long way in IRD.

\subsection{Beyond Benchmark Scores}
\label{sec:results-human-eval}

\begin{figure}[t]
    \centering
    \includegraphics[width=1\linewidth]{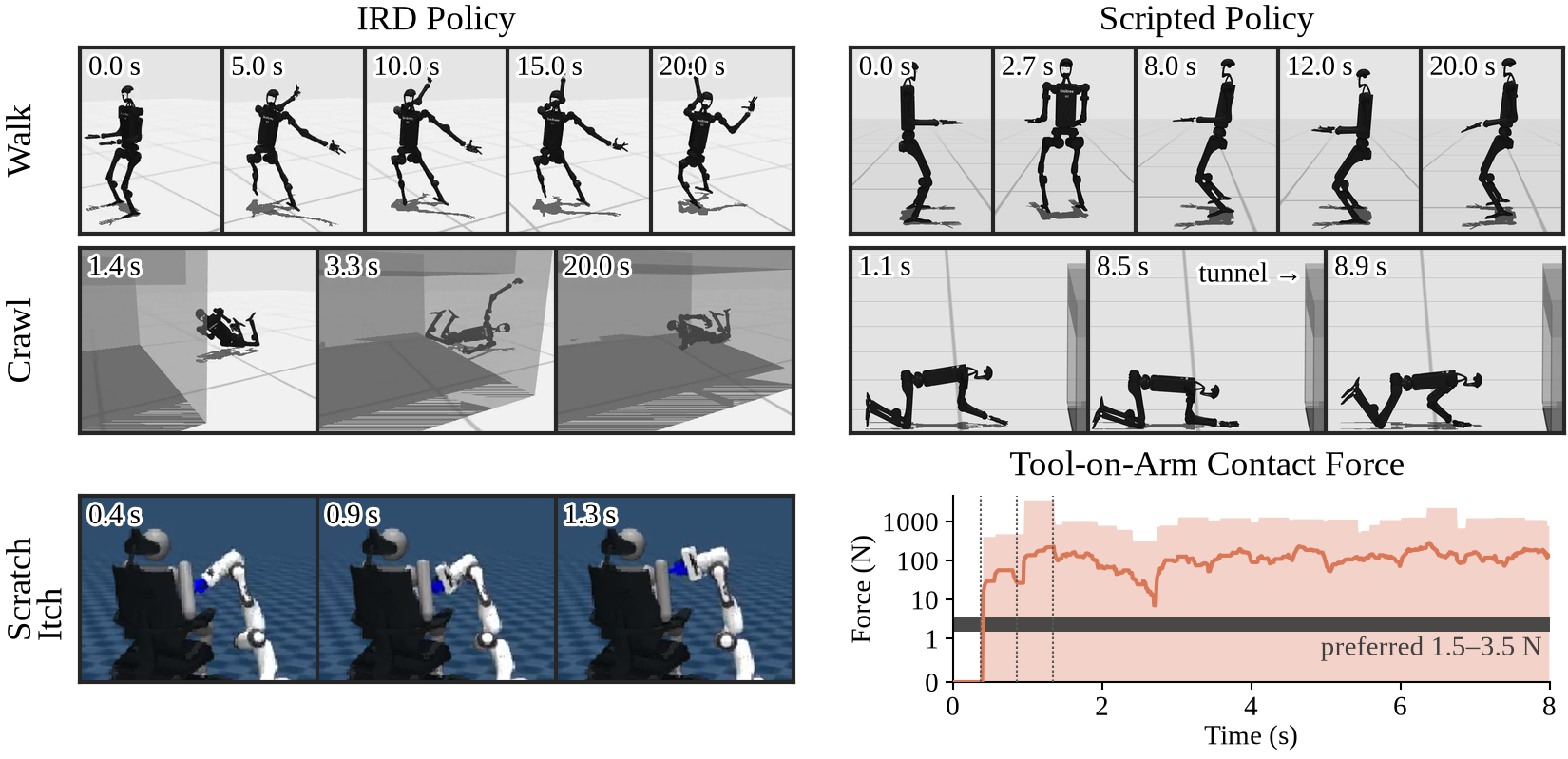}
\caption{\textbf{Examples of high success, low naturalness.} All rollouts achieve $\geq75\%$ reference reward. \textit{Top left:} IRD policies on HumanoidBench flail frenetically on walk and crawl. \textit{Top right:} scripted policies earn comparable reference reward with even stranger behavior: hopping facing backwards, rocking in place. \textit{Bottom:} The IRD scratch-itch policy scores well on the reference reward, but poorly on Assistax's human preference reward: at one point applying over 3\,kN of force to the patient's arm, enough to cause severe pain \citep{iso15066}.}
\label{fig:align-case-study}
\end{figure}

Higher benchmark scores are only useful if they reflect the behavior the designer intended. To check this, two authors rated policy rollouts for \emph{completion} (was the task achieved?) and \emph{naturalness} (was it achieved the way the rater expected?). Details of the rating process can be found in \cref{app:human-eval}. Although this is admittedly a small study, our claims are a negative result about low naturalness that we expect the majority of people would agree with. We find that none of the evaluated methods, including ERA-U and ERA-S, achieve consistent success and naturalness on Assistax and HumanoidBench despite achieving some success according to reference rewards. We formulate our hypotheses about IRD failure modes below.

\textbf{Visual verification is unreliable.} Human completion ratings track benchmark scores closely on Meta-World ($\rho = 0.78$) but only weakly on Assistax ($0.36$) and HumanoidBench ($0.26$; \cref{tab:human-benchmark}). In these domains, success depends on quantities that are hard to see, such as contact force when scratching an arm, or hard to judge visually, such as forward velocity in a poor gait. We hypothesize that judgments humans cannot make accurately from video alone will also be hard for VLM judges. Therefore, we propose that when VLM judges are used to inform the reward, it is crucial that they are provided with sufficient information to judge task success (e.g. access to velocity if task is about speed).

\begin{wraptable}{r}{0.46\linewidth}
\vspace{-\baselineskip}
\centering\small
\caption{\textbf{Humans agree on completion more than on naturalness.} Gym MuJoCo clips were rated for completion only, and HumanoidBench naturalness ratings are almost all at the minimum, so that cell is not reported, as no informative correlation can be calculated.}
\label{tab:human-benchmark}
\begin{tabular}{@{}lcc@{}}
\toprule
& Compl. & Natur. \\
\midrule
Rater A vs.\ B          & 0.83 & 0.54 \\
Human vs.\ benchmark    & 0.58 & 0.47 \\
\quad MT10              & 0.78 & 0.71 \\
\quad Gym MuJoCo        & 0.57 & --   \\
\quad Assistax          & 0.36 & 0.29 \\
\quad HumanoidBench     & 0.26 & --   \\
\bottomrule
\end{tabular}
\end{wraptable}

\textbf{Naturalness is hard to agree on.} Alignment requires both that the task is done and that it is done as humans intend. We find that correlation between the two reviewers is higher on task completion ($\rho = 0.83$) than on naturalness ($\rho = 0.54$).  We also find that reference rewards track naturalness ratings less well than they track completion ratings ($0.47$, \cref{tab:agreement-by-domain}) showing that naturalness is less closely tied to the reference reward. However, naturalness ratings on HumanoidBench are almost unanimous: 70 of 73 ratings gave a naturalness score of 0 and every one of the 73 was given with high confidence. On the manipulation domains the same question is far less settled, with high confidence on 60\% of Meta-World ratings and 47\% of Assistax ratings. This asymmetry matters for anyone proposing visual assessments of trajectories as a training signal: it is reliable where the embodiment makes failure legible, and noisy everywhere else.

\textbf{Reference rewards are hackable.} We find, consistent with prior work on specification gaming \citep{krakovna2020specification}, that HumanoidBench's walk and crawl reference rewards can be maximized without walking or crawling as a human would. Of 67 learned policies, 9 exceed 75\% reference reward, and raters judged all 9 unnatural (naturalness $\leq 2$). This could be an artifact of RL finding strange policies; we rule this out with two scripted policies that reach comparable reference reward by hopping facing backward or rocking in place (\cref{fig:align-case-study}). We elaborate on the implications of this in the discussion.

\section{Discussion}
\label{sec:discussion}

\textbf{Two small bitter lessons for reward design.}
\citet{sutton2019bitter} famously argued that general-purpose methods that scale with compute tend, in the long run, to outperform approaches built around human intuition. RL reward design offers two modest parallels. First, LLM-based search can produce rewards that outperform those written by hand \citep{ma2024eureka}. Second, a wide search can help improve the reward-design procedure itself. Several choices we expected to help did not: tapering the number of candidates per round, for example, made little difference at an equal budget (\cref{app:isoquant}). Yet a small, semi-manual hill-climb over simple design choices produced ERA-U and ERA-S (\cref{sec:results-recipe}). These findings suggest that empirical search deserves a larger role in designing IRD methods, especially as stronger models and automated experimentation make that search more effective.

\textbf{Alignment.}
Better search, however, only helps to the extent that its objective captures what we want. Higher benchmark scores did not reliably correspond to the behavior we intended (\cref{sec:results-human-eval}). Task descriptions usually specify what to achieve while leaving much of how to achieve it unstated. We hypothesize that LLM designers, as aggressive reward seekers \citep{pan2022effects, pan2024feedback}, will widen rather than close the gap between score and naturalness as IRD methods become stronger unless naturalness becomes part of the objective or feedback.

\textbf{Scalable oversight.} Models can reduce the human effort of evaluating behavior, but relying on their judgments to guide further optimization risks propagating errors that those same evaluators may not detect. In our study, a VLM judge scored clips whose camera views obscured task-relevant outcomes; human inspection exposed these problems and prompted re-rendering. We also encountered inaccurate task descriptions, illustrating that both the objective and the evidence used to assess it can be unreliable. These are problems that any IRD system would need to address: automated changes should not be accepted as improvements merely because they satisfy an evaluator whose inputs or judgments are flawed. Our observations therefore caution against trusting current model-based feedback as the sole oversight mechanism for autonomous improvement. Human review helped correct specific problems here; scaling this process requires systematic checks of whether evaluators can observe the relevant outcomes and whether their judgments track the intended goals.

\textbf{The future of IRD.} As models become more capable, we suspect that new paradigms of IRD will become possible. VLMs may be able to emit reward values directly without needing to represent them programmatically; they may also natively be able to reliably control embodiments, or write Python programs that do so.

\section{Conclusion}

We introduced BIRD, a framework that expresses published IRD methods as configurations of one loop, so that they can be compared, ablated, and recombined under matched feedback conditions and policy-training budgets. Our experiments suggest two takeaways. First, simple design choices were the most robust: in our matched setting, the benefits of many design choices, including many of our own proposals, did not carry over consistently across tasks and methods, which may reflect how strongly those benefits depend on the setting, while a few simple choices helped consistently. Second, search over such choices is productive: a small hill-climb starting from RDA and drawing on components from prior methods, such as RF-Agent's design thought, produced the two strongest methods we evaluated, and we expect further search to improve on them. Our human study, though small, also suggests a limit to this approach: hill-climbing on benchmark scores improves performance but not necessarily the intended behavior, and encoding naturalness and human intent in the objective remains a hard, open problem for the field. We describe limitations in \Cref{app:limitations}.

\newpage

\bibliography{refs}
\bibliographystyle{iclr2027_conference}

\newpage

\appendix
\makeatletter
\setlength{\@fptop}{0pt}
\setlength{\@fpsep}{\floatsep}
\setlength{\intextsep}{10pt plus 2pt minus 2pt}
\makeatother
\crefalias{section}{appendix}
\crefalias{subsection}{appendix}
\crefalias{subsubsection}{appendix}

\section{Terminology}
\label{app:terminology}

\begin{table}[htbp]
  \centering
  \small
  \setlength{\tabcolsep}{5pt}
  \renewcommand{\arraystretch}{1.18}
  \caption{Terminology used throughout the paper. Examples illustrate reward design for a cube-manipulation task.}
  \label{tab:terminology}
  \begin{tabular}{@{}
    >{\raggedright\arraybackslash}p{0.17\linewidth}
    >{\raggedright\arraybackslash}p{0.40\linewidth}
    >{\raggedright\arraybackslash}p{\dimexpr0.43\linewidth-4\tabcolsep\relax}
    @{}}
    \toprule
    \textbf{Term} & \textbf{Definition} & \textbf{Example} \\
    \midrule
    \textbf{Task objective}
    & The intended behavior the policy should achieve.
    & Pick up a cube and hold it at the target height for two seconds. \\
    \addlinespace
    \textbf{Task description}
    & The natural-language statement of the task objective given to the generator, judges, and raters.
    & ``Lift the cube and hold it at the green marker.'' \\
    \addlinespace
    \textbf{Reward function}
    & The function providing the learning signal during policy training. A \emph{candidate reward} is one proposed during search.
    & $R = R_{\mathrm{reach}} + R_{\mathrm{lift}} + R_{\mathrm{hold}}$, rewarding reaching, lifting, and maintaining the hold. \\
    \addlinespace
    \textbf{Reference reward}
    & The reward function distributed with the environment. It is withheld from the generator.
    & The benchmark's hand-designed dense reward for the cube task. \\
    \addlinespace
    \textbf{Benchmark score (fitness)}
    & The value a benchmark assigns a trained policy with respect to the task objective; formally, the fitness $F$. Reward design seeks to maximize $\mathbb{E}[F(\mathcal{A}_M(R))]$.
    & The rate of successful two-second holds: eight successes in ten evaluation episodes give $F = 0.8$. \\
    \addlinespace
    \textbf{Feedback}
    & Information used to guide reward search, including benchmark scores, model or human judgments, and training statistics.
    & A benchmark score of $0.4$, or ``The cube reaches the target height but is immediately dropped. Increase the reward for maintaining the hold.'' \\
    \addlinespace
    \textbf{Supervised / unsupervised IRD}
    & Whether a method uses the benchmark score as feedback during search (supervised) or never sees it (unsupervised).
    & Eureka selects the reward whose policy has the highest success rate (supervised); RDA asks a VLM to judge rollout frames (unsupervised). \\
    \addlinespace
    \textbf{Design choice}
    & A configurable aspect of the IRD procedure.
    & Policy initialization: \texttt{from\_scratch} or \texttt{warm\_start}. \\
    \addlinespace
    \textbf{Key}
    & Each design choice is represented as a \emph{key} in a method's configuration YAML file.
    & The default BIRD algorithm sets \texttt{checkpoint\_selection=final}. \\
    \addlinespace
    \textbf{Configuration}
    & The complete set of selected settings for the design choices that specifies an IRD procedure.
    & See \Cref{fig:bird-config}. \\
    \addlinespace
    \textbf{Policy-training budget}
    & The amount of policy training allocated to reward search, specified per training run and across the whole search.
    & $5$ iterations $\times$ $10$ candidates per iteration $\times$ $1\mathrm{M}$ environment steps per policy-training run $=$ $50\mathrm{M}$ environment steps total. \\
    \addlinespace
    \textbf{Completion rating}
    & A human or model judgment of how fully the displayed behavior achieves the task objective.
    & ``5/5: the cube is held at the target height for two seconds.'' \\
    \addlinespace
    \textbf{Naturalness rating}
    & A human judgment of how closely the displayed behavior matches their expectations about how the task objective should be achieved.
    & ``2/5: the hold succeeds, but the arm repeatedly jerks.'' \\
    \bottomrule
  \end{tabular}
\end{table}

\FloatBarrier
\section{Prior Work}
\label{app:prior-work}

\paragraph{Reward optimality.}
The history of iterative reward design began with the observation that
not all rewards are created equal. \citet{singh2009where} explored this point by distinguishing between the \emph{reward} the agent learns to maximize and the \emph{fitness} by which the designer judges the usefulness of that reward. Intuitively, the designer cares about what the agent learns to do, rather than the return it earns under a particular reward. For example, a grasping reward might combine gripper proximity, contact, and object height, but the designer judges its usefulness by how reliably the trained agent picks up an object. Formally, this is written as a \textit{reward
design problem} (RDP)
\begin{equation}
  R^\star \in \arg\max_{R \in \mathcal{R}}
  \mathbb{E}\!\left[F\!\left(\mathcal{A}_M(R)\right)\right],
  \label{eq:rdp}
\end{equation}
where $\mathcal{R}$ is the space of candidate rewards,
$\mathcal{A}_M$ is a learning procedure that trains a policy in
environment $M$ under reward $R$, and $F$ evaluates that policy.\footnote{We use the formulation and terminology from
\citet{ma2024eureka}.}

\paragraph{Early automated search.}
While \citet{singh2009where} formulated the optimal reward problem,
they remained agnostic about how best to solve it. Their experiments
used exhaustive searches, motivating subsequent work on finding useful
rewards more efficiently. Evolutionary approaches searched over symbolic
reward programs, while gradient-based approaches exploited an agent's
planning structure to improve its reward online
\citep{niekum2010genetic,sorg2010reward}. Reward optimization subsequently
developed into methods for learning intrinsic rewards alongside policy
learning \citep{zheng2018learning} and large-scale tuning of predefined
reward components, as in AutoRL \citep{faust2019evolving}. These
developments made reward search more practical, but faced different
constraints: flexible program search required many evaluations, while
more structured approaches depended on a specified reward family or
access to the agent's optimization process.

\paragraph{LLM-guided reward design.}
The advent of coding LLMs offered a new way to guide reward search
using task descriptions, environment code, and natural-language
feedback. Rather than relying only on generic search operators, coding
models could use their programming knowledge to propose and revise
programs---a possibility explored more broadly in evolutionary search
by \citet{lehman2022evolution}. Eureka \citep{ma2024eureka} applied this
idea to reward design, retaining an evolutionary loop in which candidate
rewards are evaluated through policy training and revised using feedback
on the results. Text2Reward \citep{xie2024text2reward} similarly generated
reward programs and supported iterative refinement through human
feedback. These methods automated much of the manual design loop, but
still had to decide which candidate rewards were worth training policies under and how to
judge the behavior they produced.

\paragraph{Extensions to the loop.}
With this loop in place, recent work asks how to make better use of
each policy-training run. Broader search histories and trajectory-based
screening offer ways to revisit promising directions and reject
unsuitable rewards before training again
\citep{hazra2025revolve,gao2025rfagent,sun2025card}. The reward is also
only one part of the learning setup: extending search to policy
observations and training distributions allows these choices to be
designed alongside it \citep{jaswal2026limen,ma2024dreureka}. Across these
extensions, however, better search still requires a way to judge
behavioral improvement. This has motivated VLM-based judgments of policy
rollouts as an alternative to benchmark-score feedback, allowing
task descriptions to guide candidate comparisons and subsequent
revisions \citep{lee2026rda,ma2025gran}.

We select the methods represented in BIRD to cover these different
search procedures, design spaces, and feedback assumptions.
\Cref{fig:bird-config} summarizes their configuration differences;
\Cref{app:methods} describes each represented variant and distinguishes
implementation coverage from the methods evaluated in our experiments.

\FloatBarrier
\section{Methods in BIRD}
\label{app:methods}

BIRD implements 14 published IRD methods (\cref{tab:ird-eval-select}). We briefly describe each below.\footnote{Two early methods, the reward search of \citet{singh2009where} and Language to Rewards \citep{yu2023language}, rely on exhaustive search and model-predictive control. We include them in simplified form to show that BIRD's design space extends beyond LLM-written rewards trained with RL.}

\begin{table}[tb]
  \centering
  \small
  \caption{\textbf{Methods in BIRD.}
  Feedback: S~=~score, C~=~per-component reward statistics, T~=~trajectories, N~=~text critique.
  Optimal Reward Search and Language to Rewards are omitted (no policy-training feedback loop).
  $^a$LLM-written critics refit reward parameters.
  $^b$An LLM rates reward designs as a search prior.
  $^c$The original paper's own automated variant (REvolve: the benchmark score as its hand-designed fitness; Gran Turismo: an LLM in place of human feedback).
  $^d$Shown to the generator, but never used to rank.
  $^e$Interactive variant.}
  \label{tab:ird-eval-select}
  \vspace{4pt}
  \begin{tabular*}{\linewidth}{@{\extracolsep{\fill}}ll cccc c l@{}}
    \toprule
    & & \multicolumn{4}{c}{\textbf{Evaluator}} & & \\
    \cmidrule(lr){3-6}
    Method & Ranking signal & Bench. & VLM & Human & LLM & Feedback & Selection \\
    \midrule
    \multicolumn{8}{@{}l}{\textbf{Supervised}} \\
    Eureka           & benchmark      & \yes     & \no  & \no  & \no      & S\,C    & argmax \\
    DrEureka         & benchmark      & \yes     & \no  & \no  & \no      & S\,C    & argmax \\
    ROSKA            & benchmark      & \yes     & \no  & \no  & \no      & S\,C    & elitist \\
    R$^\ast$         & benchmark      & \yes     & \no  & \no  & \yes$^a$ & S\,C    & argmax \\
    RF-Agent         & benchmark      & \yes     & \no  & \no  & \yes$^b$ & S\,C    & UCT tree search \\
    LaRes            & benchmark      & \yes     & \no  & \no  & \no      & S       & elitist \\
    LIMEN            & benchmark      & \yes     & \no  & \no  & \no      & S       & MAP-Elites \\
    REvolve$^c$      & benchmark      & \yes     & \no  & \no  & \no      & S\,C    & island EA \\
    CARD             & none           & \yes$^d$ & \no  & \no  & \no      & S\,C\,T & none \\
    \midrule
    \multicolumn{8}{@{}l}{\textbf{Unsupervised}} \\
    RDA              & VLM score      & \no      & \yes & \no  & \no      & S\,N    & argmax \\
    Gran Turismo$^c$ & LLM preference & \no      & \yes & \no  & \yes     & C\,N    & Bradley--Terry \\
    Text2Reward$^e$  & none           & \no      & \no  & \yes & \no      & N       & none \\
    \bottomrule
  \end{tabular*}
\end{table}

\textbf{Optimal Reward Search} \citet{singh2009where} argue that the best reward for a learning agent need not be the fitness its designer cares about, and define the optimal reward as the one under which the agent accumulates the most fitness. They find it by exhaustive search. Each candidate reward assigns a value from a discretized range to each combination of a few state features, such as the agent's hunger and thirst. A Q-learning agent \citep{watkins1992q} is trained under each candidate, and the reward whose agent accumulates the most fitness over its lifetime is kept.

\textbf{Language to Rewards} \citet{yu2023language} argue that LLMs struggle to output low-level robot actions directly, and use reward functions as the interface between language and control. A ``motion descriptor'' rewrites the user's instruction as a templated description of the desired motion, and a ``reward coder'' turns it into code that sets the weights of predefined reward terms. MuJoCo MPC \citep{howell2022predictive} then optimizes the reward online, so no policy is trained, and a person types corrections for the next round.

\textbf{Eureka} \citet{ma2024eureka} cast reward design as evolutionary search over reward code. Each iteration, an LLM samples a batch of reward programs from the task description and the environment's source code. A policy is trained under each, and the reward with the best task score is kept. A ``reward reflection'' summarizing each reward component's values over training then seeds the next iteration.

\textbf{Text2Reward} \citet{xie2024text2reward} generate dense reward code from a natural-language instruction, showing the LLM a hand-written Pythonic abstraction of the environment rather than its source. Execution errors are returned to the LLM until the code runs, and a few-shot variant retrieves an expert-written reward from a related task. An interactive variant revises the reward using a person's free-text feedback on rollout videos.

\textbf{DrEureka} \citet{ma2024dreureka} extend Eureka to sim-to-real transfer. The reward search adds a safety instruction asking for stable, smooth motion. After the search, a reward-aware physics prior finds the range of each physics parameter over which the policy still succeeds, and the LLM writes domain-randomization configurations within those ranges. Policies trained under them are compared on the real robot.

\textbf{REvolve} \citet{hazra2025revolve} target tasks such as driving, where good behavior is hard to quantify, and argue that refining a single winner, as Eureka does, discards useful candidates. REvolve runs an island-model evolutionary search: each offspring is produced by mutation or crossover of parents sampled by fitness from one island, and it joins that island only if its fitness is at least the island mean. Fitness is an Elo rating \citep{elo1978rating} over human preferences between rollout videos, and human critiques are added to the prompt.

\textbf{CARD} \citet{sun2025card} aim to reduce the cost of iterative reward design in human feedback, LLM queries, and RL training. A Coder writes one reward per round, and an Evaluator reports training curves, success rates, and example rollouts back to it. Their Trajectory Preference Evaluation checks whether a new reward scores stored successful trajectories above failed ones, and a reward that fails this check is not trained.

\textbf{RF-Agent} \citet{gao2025rfagent} argue that greedy and evolutionary searches make poor use of past feedback, and cast reward design as Monte Carlo tree search over trained rewards. Each round descends the tree by UCT \citep{kocsis2006bandit} to a leaf and expands it with LLM actions such as structure mutation, weight mutation, and crossover with elite rewards. The best task score is backed up the tree, and an LLM's rating of each reward's design serves as an exploration prior.

\textbf{R$^\ast$} \citet{li2025rstar} extend Eureka in two ways. First, a quarter of each batch comes from crossover that inserts a reward component from one archived reward into another, without an LLM call. Second, LLM-written critic programs compare trajectory segments, and each reward's numeric parameters are refit to these preferences with a Bradley--Terry loss \citep{bradley1952rank}.

\textbf{ROSKA} \citet{huang2025roska} argue that training each candidate's policy from scratch, as Eureka does, is inefficient. Each new candidate instead starts from a blend of the previous best policy and a random initialization, with the blend ratio chosen by Bayesian optimization \citep{snoek2012practical} over short training probes. The search also keeps sampling rewards until one beats the incumbent.

\textbf{LaRes} \citet{li2025lares} target sample efficiency by pairing each of a population of LLM-written rewards with its own off-policy agent. All agents share one replay buffer, relabeled under every member's reward, and Thompson sampling \citep{thompson1933likelihood} decides which agent collects data. Periodically, the LLM rewrites the less successful rewards from the best one, and their agents restart from the best agent.

\textbf{Gran Turismo} \citet{ma2025gran} design rewards for a racing agent without access to a fitness function during search. An LLM writes reward components from a description of the state and action space, and a filter trains only the candidates most consistent with earlier preferences. A VLM describes paired clips of the trained agents, an LLM picks the better agent in each pair, and Bradley--Terry ranking selects the winner, on which a person gives feedback.

\textbf{RDA} \citet{lee2026rda} observe that coarse numeric feedback such as success rate produces policies that reach the goal while ignoring the instruction. A VLM first decomposes the instruction into subtasks. It then scores each candidate's rollout frames against each subtask, and the reward with the best mean score is kept. Reflection revises both the reward and the subtask list, and policies can warm-start from the previous best.

\textbf{LIMEN} \citet{jaswal2026limen} note that LLM reward design assumes a fixed observation space, and instead search over the whole RL interface. Each candidate is a program that outputs both an observation map and a reward. A MAP-Elites archive \citep{mouret2015illuminating}, binned by observation size and program complexity, keeps the best interfaces and supplies parents for the next candidate.

\FloatBarrier
\section{Limitations}
\label{app:limitations}

We observe six main limitations. First, human ratings come from two authors and may not represent broader human preferences. Second, we evaluate only a subset of IRD methods,
excluding specialized approaches such as LIMEN (iterative reward and observation design)
\citep{jaswal2026limen} and Auto-MC \citep{li2024automc} (IRD for Minecraft), which may offer
additional effective design choices. Third, we do not study computational efficiency or ways
to reduce the cost of IRD. Fourth, ERA-U and ERA-S come from a deliberately small search and
are not attempts at a state-of-the-art method. Fifth, we did not systematically vary the generator or judge model, so how our conclusions depend on model choice is untested. It is also possible that some task information has leaked into the VLMs, as the benchmarks are publicly available. Finally, our evaluation covers robotic control only, leaving other RL domains untested.

\FloatBarrier
\section{Design choices in the ablation study}
\label{app:design-choices}

\begin{table}[p]
\centering
\caption{Design choices evaluated in the ablation study. Descriptions
refer to the enabled setting. Numbering matches Figure~5. $\dagger$ indicates design choice introduced in BIRD, $\ddagger$ indicates multiple key flips to facilitate a design choice}
\label{tab:design-choices}
\small
\setlength{\tabcolsep}{5pt}
\renewcommand{\arraystretch}{1.12}
\begin{tabular}{@{}r
  >{\raggedright\arraybackslash}p{0.30\linewidth}
  >{\raggedright\arraybackslash}p{\dimexpr0.70\linewidth-3em-20pt\relax}@{}}
\toprule
\textbf{No.} & \textbf{Design choice} & \textbf{Enabled setting} \\
\midrule
\multicolumn{3}{@{}l}{\textbf{Generate}}\\*
1 & Four-turn history & Use a history of four turns as generation context. \\
2 & Behavioral analysis in prompt & Include behavioral analysis in the generation prompt. \\
3 & Thought + parent + alignment$^{\ddagger}$ & Enable design thought, parent conditioning, and alignment together. \\
4 & Global-best parent & Use the highest-fitness reward found so far as the parent for generation. \\
5 & Co-evolve subtasks & Update the subtask decomposition alongside the reward. \\
6 & Prompt reset each iteration & Reset the prompt at each iteration. \\
7 & Design thought & Generate a description of the reward's design intent alongside the reward code. \\
8 & Reward-engineering tips & Include reward-engineering guidance in the generation prompt. \\
9 & Tapered candidate schedule$^{\dagger\ddagger}$ & Generate fewer candidates in later search rounds. \\
10 & Thought + parent's thought$^{\ddagger}$ & Enable design thought and include the parent's design thought as context. \\
11 & Task decomposition in prompt & Include a decomposition of the task into subtasks in the generation prompt. \\
12 & Dialogue history & Include dialogue history as generation context. \\
13 & Environment source code & Include the environment's source code in the generation prompt. \\
14 & Reward code as assistant turn & Represent reward code as an assistant turn in the conversation. \\
15 & Per-component output + feedback$^{\ddagger}$ & Enable per-component reward output together with per-component feedback. \\
\addlinespace
\multicolumn{3}{@{}l}{\textbf{Train}}\\*
16 & Peak own-reward checkpoint$^{\dagger}$ & Evaluate the policy checkpoint with the highest return under the candidate's own reward. \\
17 & Warm start from best policy$^{\ddagger}$ & Initialize policy training from the best policy found so far. \\
\addlinespace
\multicolumn{3}{@{}l}{\textbf{Evaluate}}\\*
18 & Summarized feedback & Provide evaluation feedback in summarized form. \\
19 & Task-success fitness & Use task success as the candidate's fitness. \\
20 & Fitness scalar in feedback$^{\dagger}$ & Include the scalar fitness value in feedback to the generator. \\
21 & Maximum fitness over training & Use the maximum fitness observed during policy training. \\
22 & Aliased metric name & Present the evaluation metric under an alias. \\
23 & Repeated VLM judging$^{\dagger}$ & Evaluate policy behavior through repeated calls to a vision-language model judge. \\
24 & Curve-similarity report & Include a curve-similarity report in the evaluation feedback. \\
25 & Numeric reflection & Include numeric reflection in the evaluation feedback. \\
26 & Per-component feedback & Provide feedback for individual reward components. \\
\addlinespace
\multicolumn{3}{@{}l}{\textbf{Select}}\\*
27 & Return global best & Return the highest-fitness reward found over the full search. \\
28 & Population of 16 survivors$^{\ddagger}$ & Retain a population of 16 surviving candidates. \\
29 & Select across all rounds$^{\dagger}$ & Select candidates from across all search rounds. \\
30 & Island EA (two demes)$^{\ddagger}$ & Use island-model evolutionary search with two demes. \\
\bottomrule
\end{tabular}
\end{table}

Table~\ref{tab:design-choices} lists the 30 design choices evaluated
in Figure~5, using the same numbering and grouping by stage of the iterative
reward design procedure. A dagger ($\dagger$) marks a choice introduced in
BIRD. A double dagger ($\ddagger$) marks an intervention that changes two or
three coupled configuration settings; its measured effect therefore applies
to the combined intervention.

We evaluate the effect of switching each choice on in the applicable
baseline configurations (Eureka, RDA, and CARD), excluding configurations in which the choice is inactive. The study covers nine tasks: six from Meta-World MT10 (peg-insert-side, pick-place, push, reach, window-close, window-open) and three from Gym MuJoCo (Hopper, Reacher, Swimmer), chosen so that the set spans manipulation and locomotion and includes the four MT10 tasks on which the baselines are not saturated (\cref{app:exp-hillclimb}). Each task is run with five seeds per experimental cell. Effects are expressed as Glass's~$\Delta$: the change in benchmark score divided by the baseline methods' standard deviation on
the corresponding task. Effects are pooled over applicable baselines and
tasks. In Figure~5, dots show individual tasks and diamonds show means
across tasks with 95\% BCa intervals computed over tasks; hollow diamonds
identify the choices marked with $\dagger$ below.

\section{Reward Distance Across Iterations}
\label{app:epic}
\suppressfloats[t]

To measure how the reward itself changes as IRD iterates, we compute the EPIC distance
\citep{gleave2021quantifying} between each iteration's best reward and the reference
reward. EPIC canonicalizes both rewards to remove potential shaping and compares them by
Pearson distance over a distribution of transitions: 0 means the two rewards induce the same
optimal policies, and unrelated rewards score about 0.7. We sample 1{,}024 transitions from
rollouts of a competent policy (Meta-World's scripted policies on MT10; PPO trained on the
reference reward on MuJoCo), so the distance reflects states a successful policy visits,
with 128 samples for canonicalization. We evaluate Eureka, RDA, CARD and Gran Turismo with
ten seeds and five iterations on all ten MT10 tasks and on the four Gym MuJoCo tasks whose
reference reward is published (half-cheetah, hopper, reacher and swimmer). In these runs,
policy training for all four methods used early stopping based on the benchmark score; this
affects which reward is retained at each iteration but not how EPIC is computed.

\begin{figure}[tb]
  \centering
  \includegraphics[width=\linewidth]{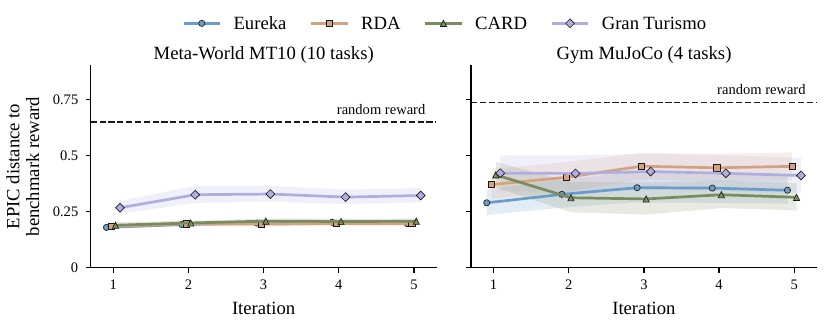}
  \caption{\textbf{Iteration does not move rewards toward the reference reward.} EPIC
  distance from each iteration's best reward to the reference reward, over transitions from
  a competent policy. Lines show means over tasks and ten seeds; bands indicate 95\%
  bootstrap confidence intervals over seeds with the task set held fixed. The dashed line
  is the median distance of random rewards, averaged over tasks.}
  \label{fig:epic}
\end{figure}

\textbf{Every method is far closer to the reference reward than chance.} On MT10, Eureka,
RDA and CARD sit at about $0.2$ and Gran Turismo at about $0.3$, against $0.65$ for random
rewards; on MuJoCo the methods lie between $0.3$ and $0.45$, against $0.74$
(\cref{fig:epic}).

\textbf{Iteration does not move the reward toward the benchmark's.} On MT10, every
method's mean distance rises slightly over five iterations, by at most $0.06$; the distance
fell significantly from the first iteration to the last in 1 of 40 method--task
combinations and rose in 10 (sign test, $p < 0.05$). On MuJoCo, Eureka and RDA drift away
($+0.06$ and $+0.08$), Gran Turismo is flat, and only CARD moves closer ($-0.10$), driven by
reacher and swimmer, where its first-iteration rewards started near chance. Closeness to the
reference reward also does not predict benchmark score: the pooled Spearman correlation between
EPIC distance and within-task benchmark-score rank is $-0.08$ on MT10 and $+0.08$ on MuJoCo, though
it reaches $\pm 0.4$ on individual tasks. The search therefore improves the trained policy
without recovering the reference reward.

\section{Candidate Allocation Isoquant Sweep}
\label{app:isoquant}
\suppressfloats[t]

A reward search spends a fixed number of candidates $C$, which can be arranged as $K$
candidates per iteration over $N = C/K$ iterations. $K = C$ is a single zero-shot round
with no feedback; $K = 1$ is $C$ sequential refinements of one reward. We ran Eureka on all
ten Meta-World MT10 tasks with $K \in \{1, 2, 4, 8, 16, 80\}$ at $KN = 80$, and with
schedules that vary $K$ across rounds at the same total. Each candidate trains a PPO policy
for 1M environment steps and is scored by the best task success over its training
checkpoints; a run's value at budget $C$ is the best such score among its first $C$
candidates. Smaller budgets are read as prefixes of the same runs (e.g.\ the first 16
iterations of a $K = 1$ run), which is valid because the number of iterations changes
nothing else in these configurations.

\begin{figure}[htb]
  \centering
  \includegraphics[width=\linewidth]{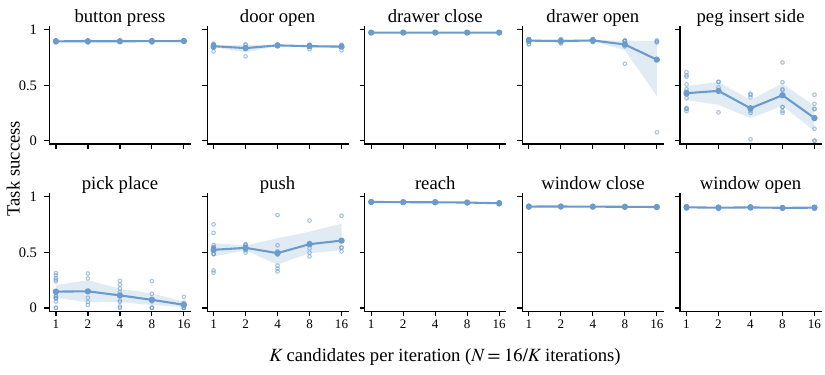}
  \caption{\textbf{Arranging a fixed budget of 16 candidates (Eureka).} Each panel is one
  MT10 task; $K$ candidates per iteration trade against $N = 16/K$ iterations, so $K = 16$
  is a single round. Lines show means over seeds (4--15 per point, median 8); hollow dots
  show individual seeds; bands indicate 95\% bootstrap confidence intervals over seeds. The
  $y$-axis is the best task success among the first 16 candidates.}
  \label{fig:isoquant}
\end{figure}

\textbf{Wide versus deep.} At $C = 16$ (\cref{fig:isoquant}), seven of the ten tasks are
flat across $K$. On drawer-open, peg-insert and pick-place, a single round of 16
candidates ($K = 16$) does worst and more iterations do better, so feedback helps where
the task is not already solved; push shows the opposite trend, within seed variation.

\textbf{Tapered schedules.} \Cref{tab:tapered-schedules} compares two schedules that vary $K$ across
five rounds, Taper (40, 20, 10, 5, 5) and Ramp (5, 5, 10, 20, 40), with constant schedules
that spend the same 80 candidates. Against $K = 8$ and $K = 80$, neither schedule differs
from the constant arm by more than $0.02$ in task success on average, and Taper and Ramp are
indistinguishable from each other ($-0.01$, 95\% CI $[-0.03, +0.01]$). Both lead $K = 16$,
but so do the other constant arms: $K = 16$ is the weakest constant arm on these seeds.
At a fixed budget, varying the number of candidates per round therefore makes little difference.

\begin{table}[tb]
\centering\small
\caption{\textbf{Tapered schedules match constant ones (Eureka, 80 candidates).} Mean
paired difference in best task success, schedule minus constant arm, paired on task and
seed, with 95\% bootstrap confidence intervals over seeds (task set fixed). The
unsaturated tasks are push, pick-place and peg-insert. Each Taper row pools 46 pairs over
all ten tasks and 12 over the unsaturated three; Ramp, 43 and 12; the constant rows, 52
and 12.}
\label{tab:tapered-schedules}
\begin{tabular}{@{}llcc@{}}
\toprule
schedule & vs.\ constant & all 10 tasks & unsaturated 3 \\
\midrule
Taper (40, 20, 10, 5, 5) & $K = 8$  & $+0.01$ $[-0.02, +0.03]$ & $+0.02$ $[-0.07, +0.10]$ \\
                         & $K = 16$ & $+0.02$ $[+0.00, +0.03]$ & $+0.06$ $[+0.01, +0.11]$ \\
                         & $K = 80$ & $+0.01$ $[-0.01, +0.02]$ & $+0.02$ $[-0.05, +0.09]$ \\
\addlinespace
Ramp (5, 5, 10, 20, 40)  & $K = 8$  & $+0.02$ $[-0.01, +0.04]$ & $+0.06$ $[-0.02, +0.13]$ \\
                         & $K = 16$ & $+0.03$ $[+0.01, +0.05]$ & $+0.10$ $[+0.03, +0.16]$ \\
                         & $K = 80$ & $+0.02$ $[+0.00, +0.03]$ & $+0.06$ $[+0.01, +0.11]$ \\
\midrule
\multicolumn{2}{@{}l}{Constant $K = 8$ vs.\ $K = 16$}  & $+0.01$ $[-0.01, +0.03]$ & $+0.05$ $[-0.04, +0.12]$ \\
\multicolumn{2}{@{}l}{Constant $K = 80$ vs.\ $K = 16$} & $+0.01$ $[-0.00, +0.02]$ & $+0.04$ $[-0.02, +0.10]$ \\
\bottomrule
\end{tabular}
\end{table}

\section{Per-Task Method Comparison Results}
\suppressfloats[t]

We provide full per-task results below.

\subsection{Performance}
\label{app:performance}

\begin{figure}[htbp]
  \centering
  \includegraphics[width=\linewidth]{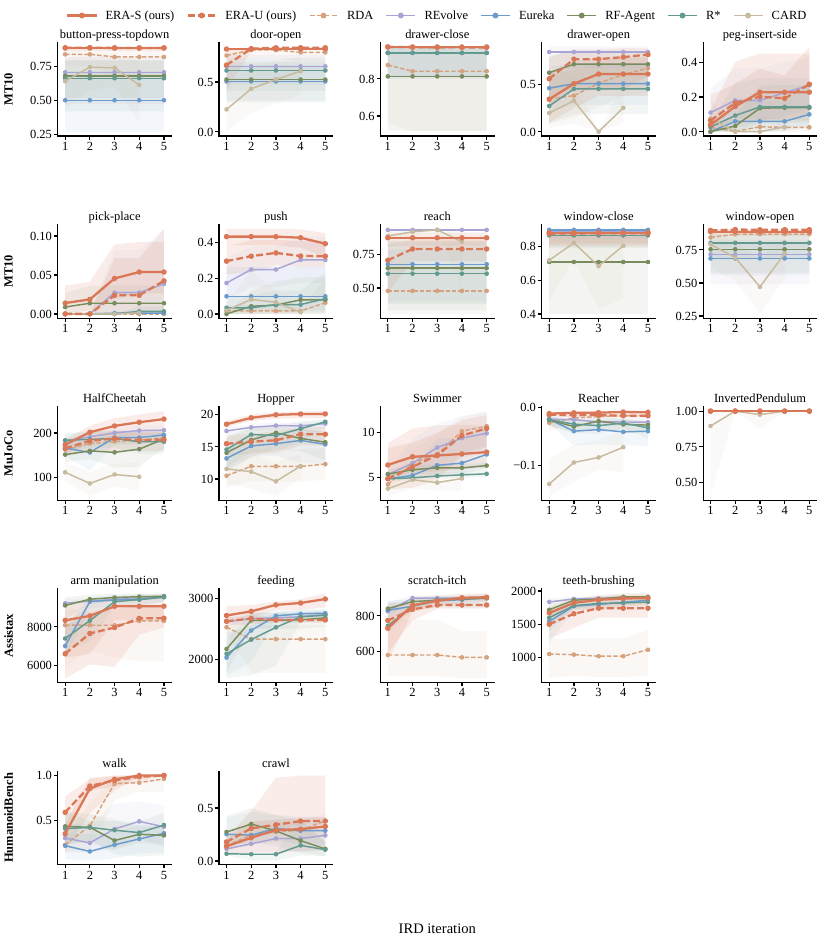}
  \caption{\textbf{Per-task results.} Benchmark score of the policy each search would return after each IRD iteration, in each task's own units: Meta-World MT10 (rows 1--2), MuJoCo (row 3), Assistax (row 4), and HumanoidBench (row 5). Lines show means over seeds (10 per task on MT10 and MuJoCo, 5 on Assistax and HumanoidBench) and shaded bands 95\% BCa bootstrap intervals. Dashed lines mark unsupervised methods. CARD, shown on MT10 and MuJoCo, trains policies in its first four iterations only.}
  \label{fig:app-all-tasks}
\end{figure}

\Cref{fig:app-all-tasks} shows every method on each of the 21 tasks. No single method is best everywhere. Excluding InvertedPendulum, where every method reaches the maximum score, ERA-S has the highest mean on 9 of the remaining 20 tasks and ERA-U on 3, the two tie on HumanoidBench walk, and four baselines are highest on the other 7 (RDA, REvolve, and RF-Agent on two each, and Eureka on one).

\subsection{Compute Efficiency}
\label{app:compute}
All evaluated methods except for CARD use near-identical environment step training budgets. CARD uses approximately 10$\times$ fewer environment steps and, as would be expected, performs worse. However, the difference in performance is not as significant as would be expected given the discrepancy in compute, suggesting that there are diminishing returns for widening the reward search space within iterations. This paper does not optimize for compute efficiency; we expect that there is a lot of potential for further improvements.

\begin{figure}[tb]
  \centering
  \includegraphics[width=0.9\linewidth]{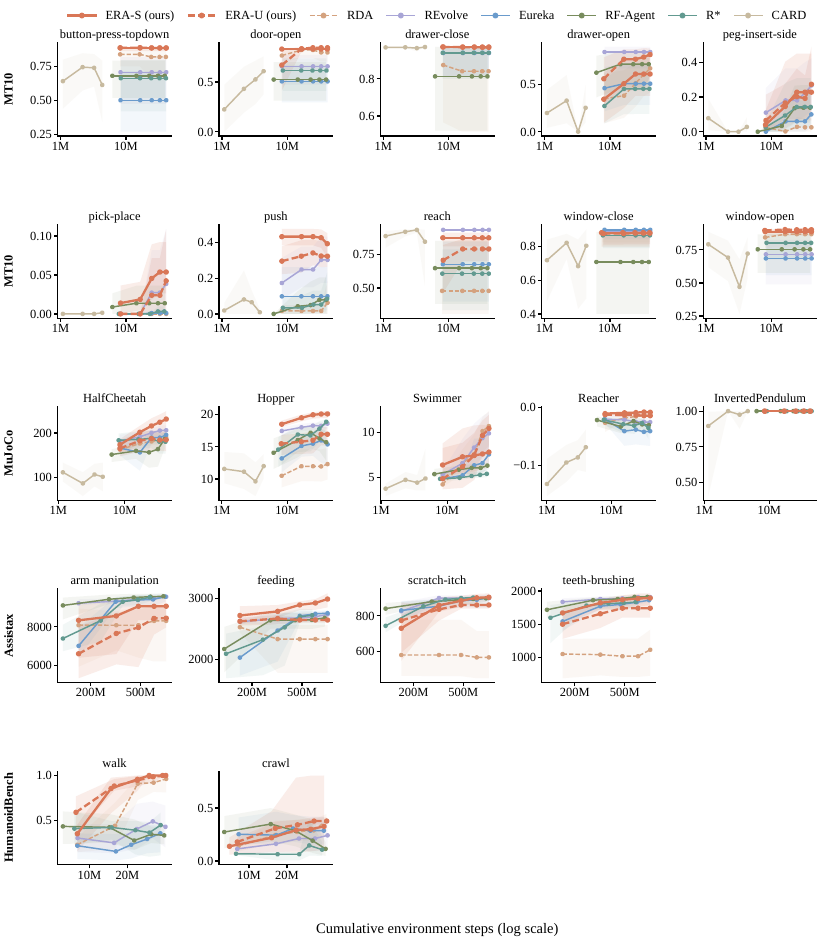}
 \caption{\textbf{Performance against environment steps.} The per-task curves of \cref{fig:app-all-tasks}, plotted against cumulative environment steps (log scale, mean over seeds). CARD trains one candidate reward per iteration and the other methods train eight, so CARD finishes with about one-tenth of their environment steps.}
  \label{fig:perf-vs-env-steps}
\end{figure}

\section{Experimental Details of IRD Runs}
\label{app:experimental-details}
\suppressfloats[t]

\subsection{Benchmarks and tasks}
\label{app:exp-tasks}

We evaluate IRD methods on 21 tasks from four benchmarks (\cref{tab:exp-tasks}). Each task is
specified once, in a task file that fixes the environment, its version, the episode horizon, the task description shown to the generator and the judges, and the \emph{reference reward}: the quantity we report for a trained policy (\cref{app:terminology}).

\begin{table}[tb]
  \centering
  \footnotesize
  \setlength{\tabcolsep}{3pt}
  \renewcommand{\arraystretch}{1.15}
  \caption{\textbf{Benchmarks and tasks.} The IRD comparison uses 21 tasks: 10 from
  Meta-World MT10, 5 from Gym MuJoCo, 4 from Assistax and 2 from HumanoidBench. Horizon is
  the episode length in environment steps; seeds is the number of independent reward
  searches per method and task. $^{\dagger}$Bed bathing is not
  in the IRD comparison, because its episodes can end early; it appears only in the human
  study (\cref{sec:results-human-eval}).}
  \label{tab:exp-tasks}
  \begin{tabular}{@{}
    >{\raggedright\arraybackslash}p{0.155\linewidth}
    >{\raggedright\arraybackslash}p{0.21\linewidth}
    >{\centering\arraybackslash}p{0.08\linewidth}
    >{\raggedright\arraybackslash}p{0.245\linewidth}
    >{\raggedright\arraybackslash}p{\dimexpr0.25\linewidth-10\tabcolsep\relax}
    >{\raggedleft\arraybackslash}p{0.06\linewidth}@{}}
    \toprule
    \textbf{Benchmark} & \textbf{Tasks} & \textbf{Horizon} & \textbf{Benchmark score}
      & \textbf{Reference reward} & \textbf{Seeds} \\
    \midrule
    Meta-World MT10, v3\newline\citep{yu2019metaworld}
      & button-press-topdown, door-open, drawer-close, drawer-open, peg-insert-side,
        pick-place, push, reach, window-close, window-open
      & 500
      & Fraction of evaluation episodes in which Meta-World's own success flag is set on
        at least one step
      & Meta-World's shipped dense reward: shaped reach, grasp and place terms, scaled to
        $[0,10]$ per step
      & 10 \\
    \addlinespace
    Gym MuJoCo, v5\newline\citep{todorov2012mujoco}
      & HalfCheetah, Hopper, Swimmer, InvertedPendulum, Reacher
      & 1000
      & A physical quantity per task (forward displacement; time balanced; final distance
        to target), normalized so that a uniform-random policy scores 0 and the best
        method's mean scores 1
      & Gymnasium's shipped reward: forward velocity less a control cost (Hopper adds an
        alive bonus); $+1$ per upright step (InvertedPendulum); negative distance to
        target less a control cost (Reacher)
      & 10 \\
    \addlinespace
    Assistax\newline\citep{hinckeldey2025assistax}
      & arm manipulation, feeding, scratch itch, teeth brushing, bed bathing$^{\dagger}$
      & 1000
      & Fraction of steps on which the tool is at its target and a contact-force condition
        holds (our metric, defined per task in the text)
      & Assistax's shipped dense reward, transcribed term by term: distance-to-target
        shaping with orientation, speed and contact-force terms, and a control cost
      & 5 \\
    \addlinespace
    HumanoidBench\newline\citep{sferrazza2024humanoidbench}
      & walk, crawl
      & 1000
      & Mean forward speed as a fraction of the 1\,m/s target, clipped to $[0,1]$; for
        crawl, multiplied by the fraction of steps spent inside the tunnel
      & HumanoidBench's shipped reward: posture terms (upright and standing for walk; low
        body height and in-tunnel for crawl) times velocity toward 1\,m/s times small
        control effort; a return of 700 counts as success
      & 5 \\
    \bottomrule
  \end{tabular}
\end{table}

\subsection{Methods compared}
\label{app:exp-methods}

\Cref{sec:results-methods} compares six published methods and our two recipes (\cref{tab:exp-methods}). Every arm is the BIRD configuration described in \cref{app:methods}, run under one execution protocol: eight candidates per iteration,
five iterations, and 1M environment steps per policy training (excluding Assistax which runs at 20M environment steps because of JAX acceleration), so a search trains 40 policies (excluding CARD). The RL optimizer is common per benchmark
(\cref{app:exp-training}) rather than each paper's own. \texttt{claude-opus-5} writes
every reward and, for RDA and ERA-U, judges the rollouts; \texttt{claude-sonnet-5} serves
the secondary evaluator roles, such as RF-Agent's self-scoring of candidate code.

\begin{table}[tb]
\caption{The compared arms. $K$ candidates per iteration, $N$ iterations (published $\rightarrow$ run here); trainings are policies trained per search. The ranking signal picks the candidate carried forward; unsupervised arms never see the reference reward or the benchmark score during search. $^{a}$CARD ranks nothing; its trajectory screen reads Meta-World's success flag and passes every reward elsewhere. $^{b}$The return of the reference reward on MuJoCo, Assistax and HumanoidBench; on Meta-World, which ships a success flag, the flag is used instead, and it coincides with the benchmark score there.}
\label{tab:exp-methods}
\begin{center}
\small
\begin{tabular}{@{}lccrl@{}}
\toprule
method & $K$ & $N$ & trainings & ranking signal \\
\midrule
\multicolumn{5}{@{}l}{\emph{supervised}} \\
Eureka \citep{ma2024eureka}      & $16 \rightarrow 8$ & 5                  & 40 & reference reward$^{b}$ \\
RF-Agent \citep{gao2025rfagent}  & 8                  & $11 \rightarrow 5$ & 40 & reference reward$^{b}$ \\
REvolve \citep{hazra2025revolve} & $16 \rightarrow 8$ & $7 \rightarrow 5$  & 40 & reference reward$^{b}$ \\
R$^{\ast}$ \citep{li2025rstar}   & $16 \rightarrow 8$ & 5                  & 40 & reference reward$^{b}$ \\
CARD \citep{sun2025card}         & 1                  & $3 \rightarrow 5$  & 4  & none$^{a}$ \\
ERA-S (ours)                     & 8                  & 5                  & 40 & reference reward$^{b}$ \\
\midrule
\multicolumn{5}{@{}l}{\emph{unsupervised}} \\
RDA \citep{lee2026rda}           & $4 \rightarrow 8$  & 5                  & 40 & VLM completion rating \\
ERA-U (ours)                     & 8                  & 5                  & 40 & VLM completion rating \\
\bottomrule
\end{tabular}
\end{center}
\end{table}

\textbf{REvolve's island count.} REvolve's paper configuration runs thirteen islands over a population of sixteen. At eight candidates per iteration, thirteen islands leave most islands with a single member. We run three islands, which keeps every island populated at this width.

\subsection{Policy training}
\label{app:exp-training}

Policy training is not modified by IRD methods tested here, so we hold it fixed and choose it for the benchmark rather than for any method: every arm's candidates train under the same RL optimizer, hyperparameters, step budget and evaluation cadence on a given benchmark.

\textbf{Meta-World and MuJoCo: PPO on CPU.} We use Stable-Baselines3 PPO
\citep{stable-baselines3, schulman2017proximal} at its library defaults.
This setup is deliberately modest. A stronger learner solves several MT10 tasks under almost any reasonable reward, which hides differences between rewards; PPO at this budget leaves room for the reward to matter.

\textbf{HumanoidBench: FastTD3 on GPU.} HumanoidBench trains with FastTD3 \citep{seo2025fasttd3} using SimbaV2 hyperspherical networks \citep{lee2025hyperspherical} at FastTD3's defaults.

\textbf{Assistax: upstream's IPPO in JAX.} Assistax trains with the benchmark's own learner rather than ours: the independent-PPO (IPPO) trainer released with Assistax \citep{hinckeldey2025assistax}, run end to end on the GPU in JAX at upstream's tuned settings. On Assistax, the whole training runs as one compiled JAX program, so for every method the policy is evaluated only once, at the end, so checkpoint-based choices (the peak own-reward checkpoint of ERA-U and ERA-S, and the maximum-over-training signals of Eureka, RF-Agent and R$^{\ast}$) have no effect there.

\textbf{Evaluation during training.} Every training is paused at up to 20 evenly spaced points to evaluate the current policy; each checkpoint records the candidate's own return, the reference reward's return and the benchmark score.

\textbf{Warm starts.} RDA, ERA-U and ERA-S initialize each candidate's policy from the best policy found so far in the search, as RDA specifies; every other arm trains from scratch. Warm starting shares parameters, not rewards: the candidate is still trained and evaluated under its own reward alone.

\subsection{Reward search protocol}
\label{app:exp-search}

A search is one run of the loop of \cref{sec:bird}: five iterations, eight candidate rewards per iteration, one policy trained per candidate under the benchmark's learner (\cref{app:exp-training}), and one reward returned at the end. This subsection fixes what every arm shares and what each arm sets; \cref{tab:exp-search} lists the per-arm settings.

\textbf{Models.} Every reward is written by \texttt{claude-opus-5}.
The same model is the vision-language judge for RDA, ERA-U and ERA-S. \texttt{claude-sonnet-5} fills the secondary evaluator roles: the text analysis behind per-subtask feedback, and RF-Agent's rating of its own candidate code.

\textbf{Verification.} Every candidate passes the same checks before training: the code parses, defines a reward function of the expected signature and uses no forbidden symbol, then runs for one environment step and returns finite values of the right shape, under a 60-second limit. What happens on failure is each method's own choice.

\textbf{Feedback and judging.} After training, each candidate is described back to the generator. Supervised arms receive the ranking signal (\cref{app:exp-methods}) together with statistics specified by their algorithm.
The unsupervised arms are judged from video. Each trained policy is rolled out three times; twenty frames, evenly spaced over the rollout, are shown to the judge.

\begin{table}[tb]
\caption{Per-arm search settings, as resolved in the reported runs. Repairs are LLM retries allowed on failed reward candidate code. Return is the reward the search hands to the final retrain.}
\label{tab:exp-search}
\begin{center}
\small
\begin{tabular}{@{}lllrl@{}}
\toprule
method & environment shown as & prompt memory & repairs & return \\
\midrule
Eureka     & full source                     & last iteration, 2 turns & 0  & global best \\
RF-Agent   & full source                     & tree path, 4 ancestors  & 8  & global best \\
REvolve    & Pythonic abstraction            & archive                 & 0  & archive best \\
R$^{\ast}$ & full source                     & last iteration          & 0  & global best \\
CARD       & Pythonic abstraction + template & full dialogue           & 10 & final incumbent \\
RDA        & full source                     & rebuilt, 4 turns        & 0  & global best \\
ERA-U/S    & full source                     & rebuilt, 4 turns        & 0  & final incumbent \\
\bottomrule
\end{tabular}
\end{center}
\end{table}

\subsection{The hill-climb behind ERA}
\label{app:exp-hillclimb}

\begin{table}[tbp]
\caption{The 24 experiments of \cref{fig:hill-climb-unsupervised}, by experiment number. $\Delta$ is the configuration's task success (best checkpoint of the returned reward) relative to RDA, averaged over pick-place, push, peg-insert-side and drawer-open. Intervals are 95\% BCa over seeds. $^{\dagger}$Fewer completed seeds on some tasks. $\bullet$ marks correspond the filled markers of the figure.}
\label{tab:hillclimb-points}
\begin{center}
\small
\setlength{\tabcolsep}{4pt}
\begin{tabularx}{\textwidth}{@{}rl>{\raggedright\arraybackslash}Xcrr@{\,}l@{\,}c@{}}
\toprule
\# & configuration & change tested & seeds & steps (M) & \multicolumn{3}{c@{}}{$\Delta$ vs.\ RDA [95\% CI]} \\
\midrule
1 & v1 & Continuous $[0,1]$ judge score; return the end of the chain; tapered candidate count at 40 trainings & 5$^{\dagger}$ & 41.3 & $+0.01$ & $[-0.09,+0.11]$ & $\bullet$ \\
2 & v2\_verify & Demonstration-based verifier before training; early stopping on own reward; 24 trainings & 3 & 10.7 & $+0.13$ & $[+0.02,+0.23]$ & $\bullet$ \\
3 & v2\_verify\_wide & Wide first round, short tail & 3 & 12.1 & $+0.12$ & $[+0.00,+0.23]$ & \\
4 & v2\_verify\_short & Two rounds only, 16 then 8 candidates & 3 & 11.9 & $-0.13$ & $[-0.22,-0.01]$ & \\
5 & v2\_verify\_adapt & Verifier sets how many passers to train & 3 & 8.0 & $+0.03$ & $[-0.15,+0.19]$ & \\
6 & v2\_verify\_kl & KL penalty toward the demonstration policy & 3 & 10.6 & $-0.14$ & $[-0.26,+0.01]$ & \\
7 & v2\_verify\_noes & Verifier without early stopping & 3 & 24.8 & $0.00$ & $[-0.12,+0.16]$ & \\
8 & v2\_verify\_es8 & Gentler early-stopping rule & 5$^{\dagger}$ & 19.4 & $-0.01$ & $[-0.14,+0.09]$ & \\
9 & v2\_verify\_es8\_constant & Control: constant width instead of the taper & 5 & 19.5 & $+0.03$ & $[-0.07,+0.13]$ & \\
10 & v2\_verify\_es8\_ramp & Control: the taper reversed & 5 & 19.7 & $-0.05$ & $[-0.18,+0.05]$ & \\
11 & v2\_verify\_es8\_distinct & Diversity by distinct per-sample prompts & 4$^{\dagger}$ & 19.5 & $-0.01$ & $[-0.15,+0.08]$ & \\
12 & v2\_verify\_es8\_personas & Diversity by system-prompt personas & 3$^{\dagger}$ & 20.1 & $-0.17$ & $[-0.31,-0.01]$ & \\
13 & v2\_noverify\_es8 & Control: verifier removed & 5$^{\dagger}$ & 19.8 & $-0.01$ & $[-0.14,+0.09]$ & \\
14 & v3\_tree & RF-Agent's search tree as persistent memory & 5 & 9.3 & $+0.08$ & $[-0.06,+0.21]$ & \\
15 & v3\_all & All five RF-Agent mechanisms together & 5 & 9.6 & $+0.10$ & $[-0.04,+0.22]$ & \\
16 & v3\_elite & Crossover with an elite set of trained rewards & 5 & 9.6 & $+0.15$ & $[+0.00,+0.28]$ & $\bullet$ \\
17 & v3\_actions & RF-Agent's distinct mutation actions & 5 & 9.6 & $+0.05$ & $[-0.10,+0.19]$ & \\
18 & v3\_puct & PUCT child selection in place of UCT & 5 & 10.4 & $+0.07$ & $[-0.06,+0.19]$ & \\
19 & v3\_selfverify & LLM self-verification score as a UCT prior & 5 & 10.2 & $-0.02$ & $[-0.16,+0.11]$ & \\
20 & v3\_thought & Design thought before the code, carried in the prompt & 5 & 11.4 & $+0.13$ & $[-0.01,+0.25]$ & \\
21 & v3\_thought\_elite & Design thought and elite crossover together & 5 & 9.6 & $+0.08$ & $[-0.07,+0.23]$ & \\
22 & v4 & Verifier removed from v3\_thought & 5 & 13.5 & $+0.01$ & $[-0.11,+0.12]$ & \\
23 & v4\_tree & Design thought on the tree-and-actions arm, verifier removed & 5 & 13.2 & $+0.01$ & $[-0.09,+0.10]$ & \\
24 & \textbf{ERA-U} & v4\_peak\_noes40: peak own-reward checkpoint; early stopping off; Eureka's 40 trainings (main-body study) & 7 & 40.0 & $\boldsymbol{+0.21}$ & $[+0.11,+0.32]$ & $\bullet$ \\
\bottomrule
\end{tabularx}
\end{center}
\end{table}

\textbf{Setup.} The climb started from RDA and ranked candidates only on the VLM
judge: no configuration saw the benchmark score or the reference reward during
search. We scored configurations by their mean task success on four MT10 tasks
(pick-place, push, peg-insert-side and drawer-open), the only MT10 tasks on which
the published methods were not saturated, with three to seven seeds each. To keep
the climb cheap, configurations 2--23 ran with 24--30 policy trainings per search,
most with early stopping on the candidate's own reward, and so used 8--25M
environment steps (\cref{tab:hillclimb-points}). Only configuration 1 and ERA-U
(configuration 24) ran the full 40 trainings of \cref{app:exp-methods}.

\textbf{Demonstrations.} Configurations 2--12 and 14--21 include a verifier that
uses rollouts of Meta-World's scripted expert policy. Before training, it checks
that a candidate reward pays the expert more than a random policy, and the rewards
it assigns to these rollouts are shown to the judge and the generator. These
configurations therefore use demonstrations, which RDA does not. The verifier
rejected none of roughly 600 candidates, and we removed it in configuration 22, so
ERA-U uses no demonstrations.

\textbf{The climb.} We ran four waves, each changing one design choice, or one
coupled group of choices, on a base taken from the previous wave. We chose bases by
judgment rather than by a fixed rule, since the intervals of the leading
configurations overlap. From the RF-Agent wave (14--21) we kept design thought
(20) rather than elite crossover (16), because the tree-based configurations
produced collapsed seeds on peg-insert-side and drawer-open. We then removed the
verifier (22) and added the peak own-reward checkpoint at the full budget, which
gave ERA-U (24). We stopped there, without a second climb, without climbing ERA-S or
any other method, and without varying the models.

\subsection{Compute}
\label{app:exp-compute}

Every search in \cref{sec:results-methods,sec:results-recipe} trains 40 policies. \Cref{tab:exp-compute} gives the cost of one search and of each benchmark's full set, measured where we have the run records and estimated from measured per-training rates otherwise.

\begin{table}[htb]
\caption{Compute per benchmark. Set totals cover the searches in \cref{fig:final-performance,fig:bt-rankings}; development runs, pilots and the hill-climb of \cref{sec:results-recipe} are excluded.}
\label{tab:exp-compute}
\begin{center}
\small
\begin{tabular}{@{}llrr@{}}
\toprule
benchmark & learner and hardware & searches & set total \\
\midrule
Meta-World MT10 & PPO (SB3), 8 vCPUs of a 128-vCPU n2d   & 800 & $\approx$4{,}200 CPU-cell-h \\
Gym MuJoCo      & PPO (SB3), 8 vCPUs of a 128-vCPU n2d   & 400 & $\approx$1{,}400 CPU-cell-h \\
Assistax        & IPPO, 1 H200 per training & 214 & $\approx$1{,}600 GPU-h \\
HumanoidBench   & FastTD3 + SimbaV2, 1 H200 per training & 93  & $\approx$2{,}200 GPU-h \\
\bottomrule
\end{tabular}
\end{center}
\end{table}

\textbf{Language-model usage.} Across the approximately 1500 IRD runs for the main results, the generator and judge together made approximately 200k model calls and consumed about 2.2 billion tokens.

\section{Human Annotation Study}
\label{app:human-eval}
\suppressfloats[t]

\subsection{Evaluation Setup}
\label{app:human-eval-setup}
\begin{figure}[htb]
    \centering
    \includegraphics[width=0.6\linewidth]{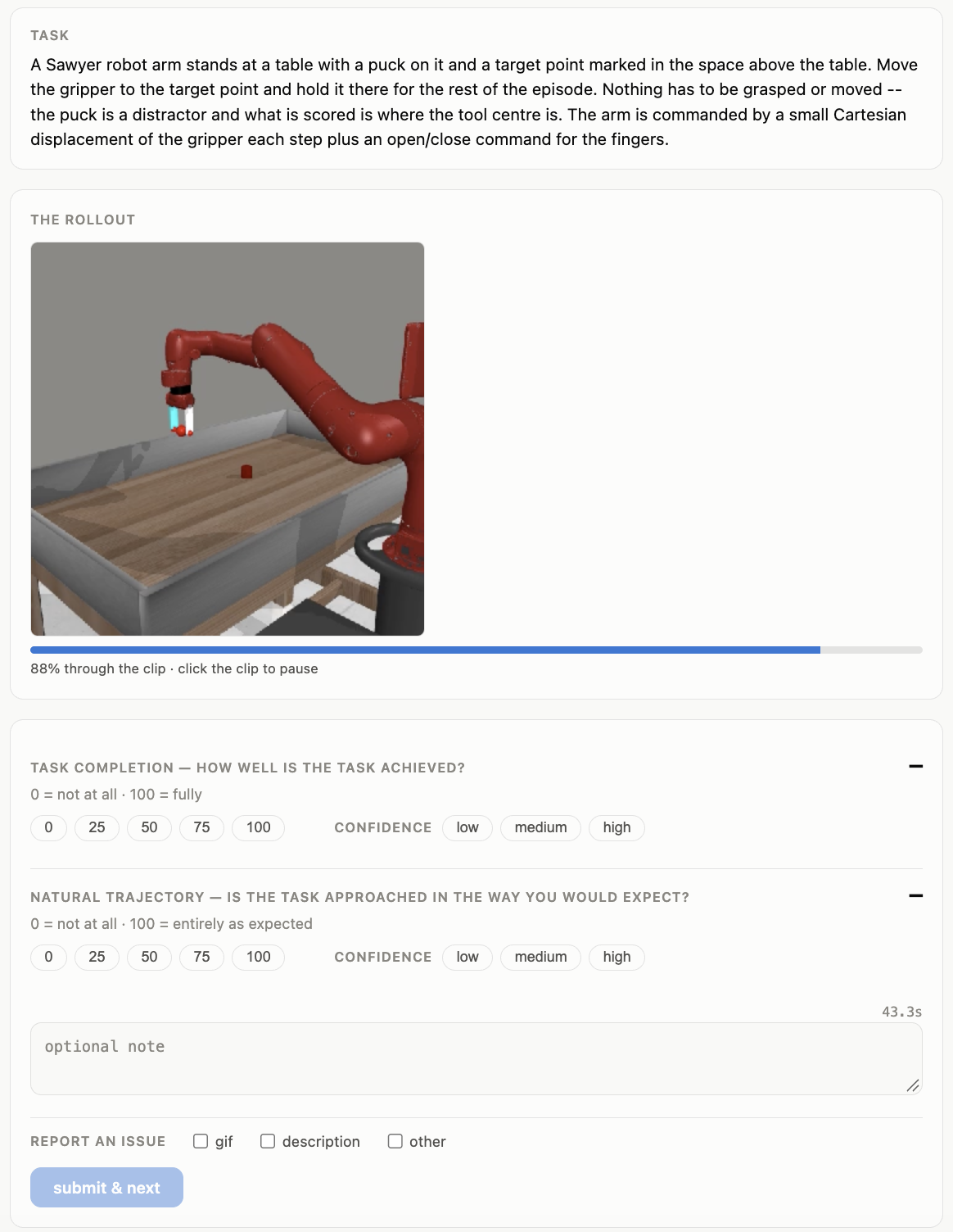}
    \caption{Screenshot of the annotation UI showing an MT10 task}
    \label{fig:annotation-ui}
\end{figure}

We collected human ratings of 257 unique episode clips across 22 tasks in Meta-World MT10, HumanoidBench, Assistax and Gym MuJoCo. Every clip is a single training rollout of a learned policy produced by one of the IRD methods under comparison, and carries the benchmark score measured on that same filmed episode. Clips were selected to cover a range of benchmark scores within each task. Two authors provided 300 usable annotations (282 first ratings and 18 repeat showings; a further 13 records were flagged by the rater as unjudgeable and excluded). Raters saw a clip and its task description, rated completion and naturalness on five-point scales, and reported confidence in each on a three-point scale. Clips were presented in randomized order across tasks and domains, and their source was hidden. Raters were blind to a clip's source but not to the study's purpose. Both raters are authors of this paper, so they are more familiar with these tasks than a typical member of the population. Analysis of the rater reliability is shown in \cref{tab:reliability}. This shows that the raters were consistent when considering the continuous nature of the rating scales, but have lower agreement when considering ratings as ordinal categories. \Cref{fig:annotation-ui} shows the interface. Raters can pause a clip and see its progress; they cannot rewind, and clips loop, which encourages repeated viewing rather than scrubbing.

\begin{table}[htb]
\caption{Rater reliability. Within-rater figures use unannounced repeat showings; between-rater figures use the clips both raters saw. $\alpha$ is Krippendorff's coefficient with an ordinal metric \citep{hayes2007answering}, which measures absolute agreement and therefore charges a constant difference in scale use as disagreement; Spearman's $\rho$ is invariant to it. Intervals are percentile bootstrap, 2000 resamples.}
\label{tab:reliability}
\begin{center}
\small
\begin{tabular}{@{}llrcccc@{}}
\toprule
comparison & question & pairs & $\alpha$ & $\rho$ & exact match & mean gap (scale steps) \\
\midrule
within rater & completion & 19 & $0.95$ & $0.91$ & 84\% & 0.16 \\
within rater & naturalness & 16 & $0.85$ & $0.71$ & 62\% & 0.38 \\
\midrule
between raters & completion & 24 & $0.74$ & $0.83$ & 42\% & 0.58 \\
between raters & naturalness & 17 & $0.57$ & $0.54$ & 53\% & 0.49 \\
\bottomrule
\end{tabular}
\end{center}
\end{table}

\subsection{Clips and sample composition}
\label{app:human-eval-clips}
\begin{figure}[htb]
    \centering
    \includegraphics[width=0.75\linewidth]{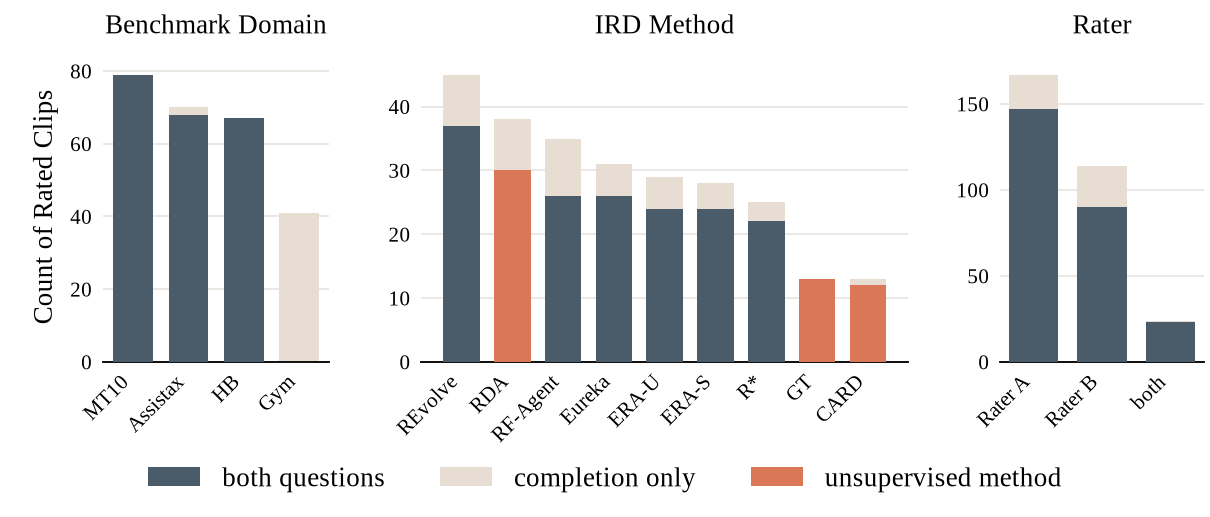}
    \caption{Breakdown of the composition of annotation data.}
    \label{fig:annotation-composition}
\end{figure}

The rated sample comprises 257 unique episode clips across 21 tasks, as shown in \cref{fig:annotation-composition}.

\subsection{Additional results analysis}
Agreement varies sharply by domain (\cref{tab:agreement-by-domain}). Agreement is highest in MT10 tasks where the tasks are more simple and easier to judge visually while Gym MuJoCo is harder due to the difficulty of visually evaluating metrics such as velocity.
On Assistax, human completion ratings are higher than benchmark scores: the arm reaches the person and holds a plausible pose, and whether the tool contacts the target is often not resolvable at the available camera distance. The opposite is true for HumanoidBench where the reference reward pays for upright, forward-moving behavior that raters do not regard as walking or crawling.

\begin{table}[htb]
\caption{Rank agreement between human completion ratings and the benchmark score, by benchmark domain. The benchmark score is the environment's own measure on the filmed episode, rescaled to $[0,1]$; on HumanoidBench it is the shipped HumanoidBench reward of the filmed episode over the 700-return success bar, clipped. Intervals are 95\% percentile bootstrap over 2000 resamples of tasks. $^{\dagger}$HumanoidBench has only two tasks, too few to resample, so its interval resamples clips and is correspondingly optimistic. The two naturalness rows pool the clips on which the second question was asked; Gym MuJoCo is completion-only by design.}
\label{tab:agreement-by-domain}
\begin{center}
\small
\begin{tabular}{@{}lrrcccc@{}}
\toprule
domain & clips & tasks & mean benchmark & mean human & $\rho$ & 95\% CI \\
\midrule
Meta-World MT10 & 79 & 10 & 0.67 & 0.69 & $+0.78$ & $[+0.60, +0.89]$ \\
Gym MuJoCo & 41 & 5 & 0.70 & 0.64 & $+0.57$ & $[+0.38, +0.72]$ \\
Assistax & 70 & 5 & 0.24 & 0.45 & $+0.36$ & $[0.00, +0.76]$ \\
HumanoidBench$^{\dagger}$ & 67 & 2 & 0.33 & 0.10 & $+0.26$ & $[+0.04, +0.47]$ \\
\midrule
all domains & 257 & 22 & 0.47 & 0.46 & $+0.58$ & $[+0.37, +0.76]$ \\
\midrule
\multicolumn{7}{@{}l}{\emph{naturalness, pooled}} \\
\quad vs.\ benchmark score & 214 & 17 & & & $+0.47$ & $[+0.17, +0.70]$ \\
\quad vs.\ human completion & 214 & 17 & & & $+0.83$ & $[+0.73, +0.87]$ \\
\bottomrule
\end{tabular}
\end{center}
\end{table}

\begin{figure}[htb]
    \centering
    \begin{minipage}[t]{0.48\linewidth}
        \centering
        \includegraphics[width=\linewidth]{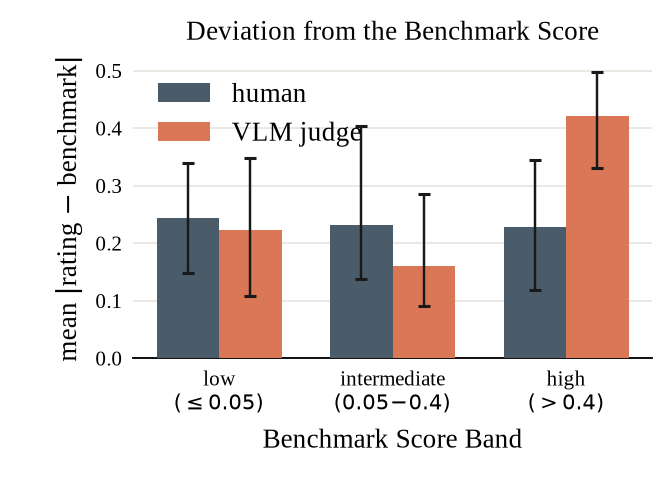}
        \caption{Mean absolute deviation between completion rating and
        benchmark score, by benchmark-score band, for human raters and the
        VLM judge (Opus 5). Lower is better.}
        \label{fig:agreement-by-score}
    \end{minipage}\hfill
    \begin{minipage}[t]{0.48\linewidth}
        \centering
        \includegraphics[width=\linewidth]{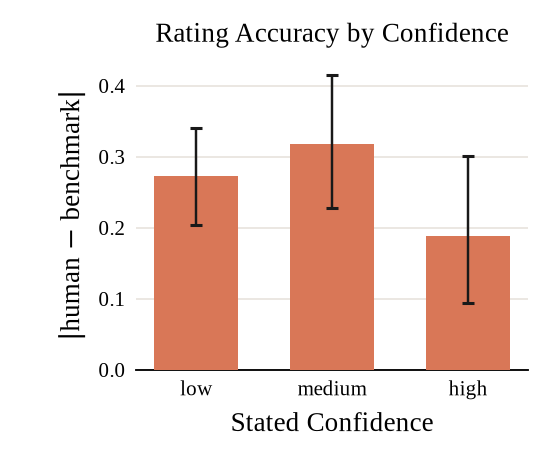}
        \caption{Rating accuracy by stated confidence: mean
        $|\text{human completion} - \text{benchmark score}|$ per rating,
        with 95\% percentile bootstrap confidence intervals. Lower is better.}
        \label{fig:accuracy-by-confidence}
    \end{minipage}
\end{figure}

Human completion ratings track the benchmark score about equally well at
every level of performance: their mean deviation stays near $0.23$ in all
three bands (\cref{fig:agreement-by-score}). The VLM judge (Opus 5) matches
humans on low-scoring policies and is slightly closer on intermediate ones.
On high-scoring policies, however, its deviation almost doubles to about
$0.42$. So the judge is least reliable where it matters most: telling a good
policy from a very good one. For unsupervised IRD methods, this means the
ranking signal gets weaker as the trained policies get better, which could
put a ceiling on the performance they can reach.

Stated confidence is at best a partial guide to accuracy. On average,
high-confidence ratings are closest to the benchmark score (mean deviation
about $0.19$). The relationship is not monotonic, though: low-confidence
ratings ($0.27$) are slightly more accurate than medium-confidence ones
($0.32$; \cref{fig:accuracy-by-confidence}). All of these differences are
tentative. The confidence intervals overlap substantially: the
high-confidence interval (roughly $0.09$--$0.30$) covers most of the
low-confidence one, so no pairwise difference is clearly resolved at this
sample size. With that caveat, the pattern still matters for human review as
an oversight mechanism. Reviewing only the low-confidence ratings would not
work, because they are no worse than the medium-confidence ones. The data
point, weakly, to a different rule: high-confidence ratings may be trusted
somewhat more, so review effort is better spent on every rating the rater did
not mark as high-confidence than on the low-confidence tail alone.

\subsection{Statistical analysis}
\label{app:human-eval-analysis}
Cross-measure agreement is summarized by Spearman's $\rho$ over clip-level scores, a clip's human score being the mean of its usable ratings. Confidence intervals are percentile bootstrap intervals from 2000 resamples of tasks, since clips within a task share an environment, a metric and often a search run; HumanoidBench contributes only two task clusters, so its intervals resample clips and are reported as such. Records carrying a rater's issue flag (13) and repeat showings (18) are excluded from the primary analyses and used only for the reliability statistics. The pooled comparison uses 257 clips across 22 tasks; the naturalness analyses use the 214 clips on which the second question was asked. High completion and low naturalness denote normalized clip means of at least $0.75$ and at most $0.25$ respectively: 75 clips are high-completion and 11 of those (15\%) are also low-naturalness. Reliability coefficients use Krippendorff's $\alpha$ with an ordinal metric, which accommodates unordered pairs and the unequal numbers of repeat showings per rater.

\section{Bradley--Terry Rankings Without the Development Tasks}
\label{app:exp-heldout}
\suppressfloats[t]

\begin{figure}[htb]
  \centering
  \includegraphics[width=\linewidth]{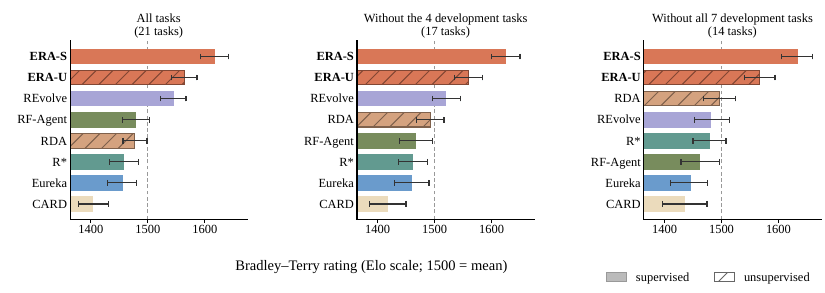}
  \caption{\textbf{Holding out the development tasks.} Pooled Bradley--Terry ratings over all 21 tasks (left), without the four MT10 tasks ERA-U was developed on (drawer-open, peg-insert-side, pick-place, push; middle), and without all seven tasks used at any point during development, which adds MT10 reach, HalfCheetah, and Swimmer (right).}
  \label{fig:bt-heldout}
\end{figure}

To verify our results, we also include the Bradley--Terry ranking on the tasks excluding the development tasks (\cref{fig:bt-heldout}). We find that ERA-S and ERA-U still rank first and second, as in \cref{sec:results-recipe}.

\section{AI Disclosure}

We used coding agents (mainly \texttt{claude-opus-5} and \texttt{claude-fable-5}) to help implement methods, polish writing, and brainstorm ideas. We also used them to help propose and refine hypotheses, design and provide feedback on research methodology and experiments, support data analysis, and interpret results. We have reviewed all AI-assisted work: AI-generated code was reviewed by the authors, and all cited literature was checked against the original papers. We take responsibility for the final content of this work, including text, claims and artifacts produced with the aid of generative AI.

\end{document}